\documentclass[sigconf]{acmart} 

\usepackage{multirow, makecell}			
\usepackage{booktabs}					
\usepackage{tabularx}					
\usepackage{threeparttable}				
\usepackage{siunitx}
\AtBeginDocument{%
  \providecommand\BibTeX{{%
    \normalfont B\kern-0.5em{\scshape i\kern-0.25em b}\kern-0.8em\TeX}}}

\copyrightyear{2021}
\acmYear{2021}
\setcopyright{acmcopyright}\acmConference[MM '21]{Proceedings of the 29th ACM
International Conference on Multimedia}{October 20--24, 2021}{Virtual Event, China}
\acmBooktitle{Proceedings of the 29th ACM International Conference on Multimedia
(MM '21), October 20--24, 2021, Virtual Event, China}
\acmPrice{15.00}
\acmDOI{10.1145/3474085.3475208}
\acmISBN{978-1-4503-8651-7/21/10}

\begin{document}
\fancyhead{}

\title{Anchor-free 3D Single Stage Detector with Mask-Guided Attention for Point Cloud}

\author{Jiale Li}
\email{jialeli@zju.edu.cn}
\affiliation{%
 \institution{Zhejiang University}
 \city{Hangzhou}
 \country{China}}

\author{Hang Dai}
\authornote{Corresponding authors.}
\email{hang.dai@mbzuai.ac.ae}
\affiliation{%
  \institution{Mohamed bin Zayed University of Artificial Intelligence}
  \city{Abu Dhabi}
  \country{United Arab Emirates}}

\author{Ling Shao}
\email{ling.shao@ieee.org}
\affiliation{%
  \institution{Inception Institute of Artificial Intelligence}
  \city{Abu Dhabi}
  \country{United Arab Emirates}
}

\author{Yong Ding}
\authornotemark[1]
\email{dingy@vlsi.zju.edu.cn}
\affiliation{%
 \institution{Zhejiang University}
 \city{Hangzhou}
 \country{China}}

\renewcommand{\shortauthors}{Li and Dai, et al.}

\begin{abstract}
Most of the existing single-stage and two-stage 3D object detectors are anchor-based methods, while the efficient but challenging anchor-free single-stage 3D object detection is not well investigated. Recent studies on 2D object detection show that the anchor-free methods also are of great potential. However, the unordered and sparse properties of point clouds prevent us from directly leveraging the advanced 2D methods on 3D point clouds. We overcome this by converting the voxel-based sparse 3D feature volumes into the sparse 2D feature maps. We propose an attentive module to fit the sparse feature maps to dense mostly on the object regions through the deformable convolution tower and the supervised mask-guided attention. By directly regressing the 3D bounding box from the enhanced and dense feature maps, we construct a novel single-stage 3D detector for point clouds in an anchor-free manner. We propose an IoU-based detection confidence re-calibration scheme to improve the correlation between the detection confidence score and the accuracy of the bounding box regression. Our code is publicly available at \url{https://github.com/jialeli1/MGAF-3DSSD}.
\end{abstract}



\maketitle

\section{Introduction}
Ongoing development of various applications such as autonomous driving \cite{KITTIDataset, luo2020c4av, dai2020commands} and robotics \cite{PC_ROBOTS} has led to the increasing demand for 3D object detectors that predict the class label, 3D bounding box and detection confidence score for each instance in a given 3D scene. In recent years, the 2D object detection methods have made great breakthroughs \cite{Maskrcnn, DCN, FCOS, Cornernet, Reppoints, meta_rcnn_acmmm20}, far ahead of the 3D object detection methods, in particular, the anchor-free detection methods. As one of the vital media for 3D space modeling and 3D visual processing, point clouds have received increasing attention from the multimedia community \cite{learningfrom3d_acmmm19, seg_pc_acmmm20, seg_pc_acmmm19, otracking_acmmm20}. The non-uniform and sparse points in the point cloud are much different from the pixels uniformly and regularly distributed in the image. So it is difficult to directly leverage the advanced 2D object detection methods for 3D object detection. Although we can convert the point cloud into an image-like 2D data structure at the input level \cite{MV3D, AVOD, WeaklySup_PC_3DOD_acmmm20, rang_img_pointcloud}, this causes the issue of losing important information such as the 3D shape of the object in the data transformation. Thus, we propose a lossless feature-level 3D-to-2D transformation. The voxel-based methods \cite{VoxelNet, Second} are a family of methods that apply the sparse 3D convolution on the voxelized point cloud in 3D space to learn the structured 3D feature volumes, which explicitly explore the 3D shape information. Since there is no perspective distortion and overlaps among objects in the Bird’s Eye View (BEV) \cite{MV3D, AVOD}, we can losslessly convert the structured sparse 3D feature volumes into the sparse 2D feature maps by flatting the features in the height dimension into the channel dimension \cite{SASSD, PVRCNN}.

\begin{figure}[!tbp]
	\centering
	\includegraphics[width=\linewidth]{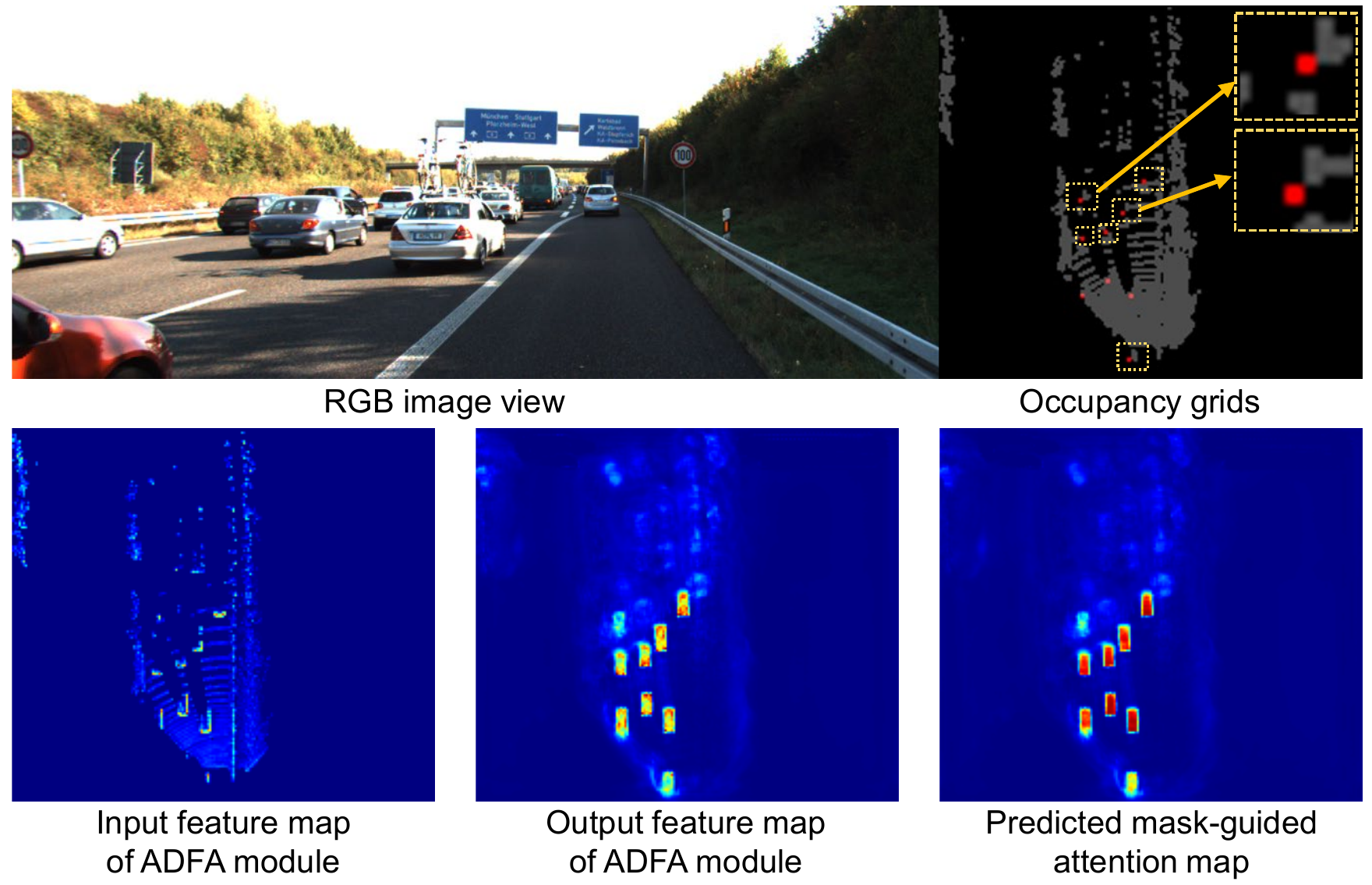}%
	\vspace{-3mm}
	\caption{Top: the RGB image and the occupancy grids of the sparse 3D feature volume in the BEV. The grid with the black pixel is empty and the red dot represents the center point of an object. Object instances with the empty central areas are highlighted by yellow boxes. Bottom: the input feature map, the output feature map of the ADFA module and the predicted mask-guided attention map. Note that the output feature map of the ADFA is much denser.}
	\vspace{-3mm}
	\label{empty_at_center}
\end{figure}

Although the sparse 2D feature maps can be generated by voxel-based feature learning, it is still far away from regressing the bounding box without anchors \cite{FCOS,Centernet}. Only the partial surface of the object is captured in the point cloud acquisition process, and most regions in the point cloud are empty. Thus, the central area of the object is also empty in the sparse 2D feature maps, as shown in Fig.~\ref{empty_at_center}. However, the 2D approaches \cite{ATSS,Centernet,FCOS} demonstrate that the central area of the object is very important in locating the object without anchors. To overcome this, we propose an Attentive Deformable Feature Adaptation (ADFA) module to make the sparse 2D feature maps fit to dense, especially in the object regions. The ADFA module incorporates the feature maps on three scales with the small receptive fields on the high-resolution features and the large receptive fields on the low-resolution features. At each scale, we employ the deformable convolution to adaptively adjust the receptive fields so that the center of the object can better perceive the entire shape information of the object. The multi-scale feature maps are further penalized with supervised mask-guided attention for highlighting the object from the complex background. 

Different from most of the existing 3D detectors \cite{Second, STD, PVRCNN}, we propose a simple and effective voxel-based 3D single-stage object detector in an anchor-free manner, which detects 3D objects as the centers of the objects in the BEV. The anchor-based 3D detectors are very popular. To handle the variable objects, they tile numerous anchor boxes of predefined locations, sizes, and even orientations in 3D space. The large searching space and the numerous anchor box attributes magnify the issues caused by the imbalance of positive and negative samples. And the anchor box attributes need to be defined manually. However, only a few methods focus on the challenging anchor-free 3D object detection \cite{Hotspots(ECCV2020)}.

The classification score from the detection head is usually used as the detection confidence assigned to the best prediction results. However, the classification score is not positively related to the location accuracy of the predicted bounding boxes \cite{IoUNet}. Thus, we propose a novel detection confidence re-calibration scheme to improve the correlation. Since the 3D Intersection-over-Union (IoU) reflects the location accuracy, a confidence estimation head is designed in parallel with other detection heads using the 3D IoUs of the predicted bounding boxes as the training target. In the inference, the predicted IoU is used to re-calibrate the classification score based detection confidence. The experimental results show that the re-calibrated confidence improves the correlation between the detection confidence and the location accuracy, and significantly enhances the 3D object detection performance.


\section{Related Work}

\textbf{Anchor-based detector.} 
As predefined anchors ease the detection process, both the 2D \cite{fasterrcnn, Maskrcnn} and 3D \cite{luo2021m3dssd,STD, PointGNN, SASSD, PVRCNN} detectors are dominated by anchor-based methods. They first evenly place numerous predefined anchors on the detection space as all potential candidates. Then, they detect objects by predicting the classes and refining the positions of anchors in one \cite{ssd, RetinaNet, VoxelNet} or several \cite{FPN, Cascadercnn,PartA2_TPAMI} times. The complex anchor boxes with different attributes need to be manually designed and tuned according to the dataset. 
 
\begin{figure*}[!tp]
	\centering
	\includegraphics[width=\linewidth]{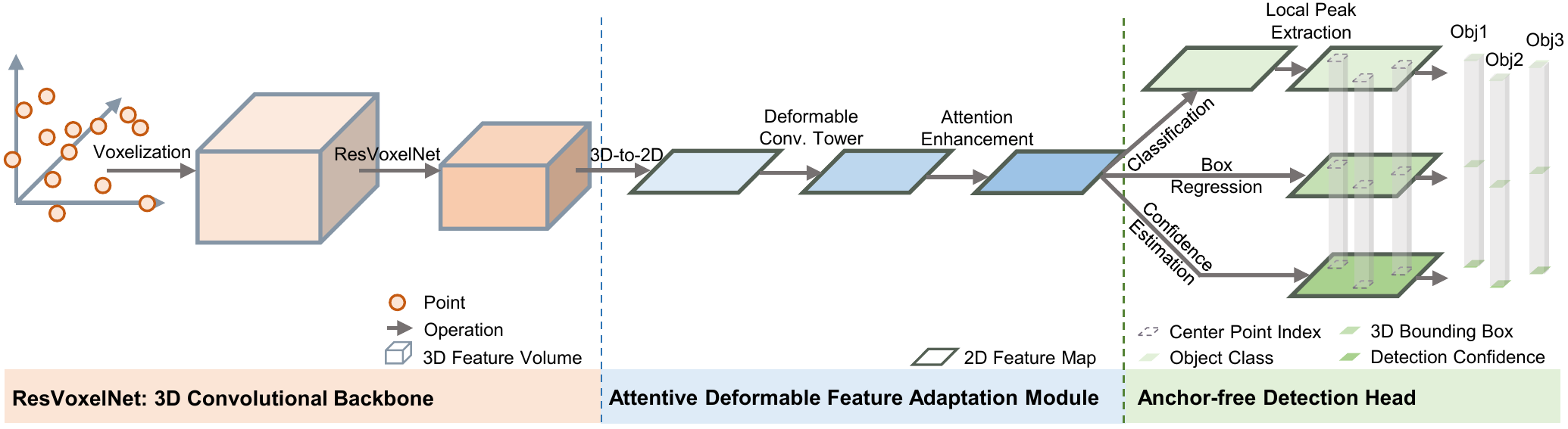}%
	\caption{The voxelized point cloud first goes through the ResVoxelNet to produce structured sparse 3D feature volume. Next, the sparse 3D feature volume is reshaped to the sparse 2D feature map in the BEV and adapted to dense by the ADFA module. Finally, three detection heads perform the center point classification, the 3D bounding box regression, and the IoU-based detection confidence estimation, respectively. The ``Obj'' is short for ``Object'' and the feature channels are not shown in this figure for clarity. Each stage can be clearly distinguished in different series of colors.}
	\label{overall_arch}
	\vspace{-2mm}
\end{figure*}

\textbf{Anchor-free detector.} 
The 2D anchor-free approaches can be grouped into two categories. One way is to represent the object as the center point or central area. In \cite{Centernet,YOLO}, only the grid cell that contains the object center is responsible for detecting this object. With the center sampling strategy \cite{ATSS}, FCOS \cite{FCOS} can be improved by regressing the box from the central area instead of the entire area inside the object bounding box, which indicates that the central area of the object can produce better predictions. The other way is to generate an axis-aligned 2D bounding box by grouping multiple defined keypoints \cite{Cornernet, CornerNet_triplets} or self-learned points \cite{ExtremeNet, Reppoints}. However, it is not feasible to transform multiple keypoints into a rotated 3D bounding box without other constraints. Recently, the anchor-free 3D detector HotSpotNet \cite{Hotspots(ECCV2020)} defines hotspots as the non-empty voxels that only lie on the surface of the object. The locations of the non-empty voxels usually vary due to the perspective issues and lack consistency compared with the centers of the objects. 

\textbf{Structured representations.}
Many methods convert unstructured point clouds to regular grids. MV3D \cite{MV3D} and AVOD \cite{AVOD} project point clouds as the 2D images in the BEV or Front View (FV). The projected 2D images can be applied to 2D convolutional networks at the cost of information loss in the projection process. VoxelNet \cite{VoxelNet} evenly divides a point cloud into small 3D voxels in \textit{XYZ} directions for 3D convolutional processing. PointPillars \cite{PointPillars} roughly generate pillars along the height direction and the following works \cite{HVNet, tanet_AAAI, li2021p2v} develop better strategies for expressive feature learning. Considering that most of the voxels are empty, SECOND \cite{Second} introduces the sparse 3D convolution \cite{sparseconv,submanifold} for efficient voxel processing. With sparse 3D convolution, these voxel-based methods \cite{PartA2_TPAMI, PVRCNN, SASSD} are efficient for accurate 3D object detection using the structured point cloud features.

\textbf{Unstructured representations.}
The PointNet \cite{PointNet} and PointNet++ \cite{PointNet++} introduce a new way of learning point-wise features directly on the point cloud without any representational transforms. F-PointNet \cite{F-PointNets} first employs \cite{PointNet++} to extract the local point cloud features in the frustum proposals generated by the image. To avoid perspective distortion and object occlusion in the image, PointRCNN \cite{PointRCNN} and STD \cite{STD} directly generate a 3D object proposal on each point of the entire point cloud \cite{li20203d}. However, the grouping of points on a large scale leads to computational costs. 3DSSD \cite{3DSSD} limited the proposals in a sampled point set, but objects may not be retained in the uneven down-sampling process. PV-RCNN \cite{PVRCNN} proposes a Voxel Set Abstraction operation to aggregate the voxel-wise features in some sampled keypoints. The two-stage methods STD and PV-RCNN both learn the IoUs of the proposal boxes to guided the NMS with detection performance improvements. Besides, Point-GNN \cite{PointGNN} directly builds a graph on the entire point cloud, and designs a graph neural network (GNN) \cite{GNN_survey} to learn vertex-wise features with local relationships, and conducts 3D object detection on vertexes. The flexible point cloud feature learning based on the points and graphs still does not change the unordered and sparse properties of point clouds.

\section{Method}
The overall architecture is presented in Fig.~\ref{overall_arch}, including three main components: 1) ResVoxelNet backbone that efficiently encodes the point cloud into the spare 3D feature volume; 2) ADFA module that conducts feature enhancement on the reshaped 2D feature map; 3) our anchor-free detection head.    
\subsection{ResVoxelNet: 3D Convolutional Backbone}
Following \cite{PVRCNN,Second,SASSD}, a point cloud is uniformly divided into 3D voxels with spatial resolution of $(L_{0}, W_{0}, H_{0})$ by a quantization step $d=(d_x, d_y, d_z)$. The feature of each non-empty voxel is initialized to the mean of the commonly used point-wise features (e.g., 3D coordinates and reflection intensity) of the points inside. The ResVoxelNet is used for the hierarchical voxel-based feature extraction, which is the refined VoxelNet \cite{VoxelNet} with residual blocks \cite{ResNet}. The network employs a series of $3\times 3 \times 3$ 3D sparse convolutional blocks to gradually convert the point clouds into feature volumes with $1\times$, $2\times$, $4\times$, $8\times$ downsampled sizes. The first layer of each block downsamples the feature volumes implemented by a normal sparse convolution \cite{sparseconv}, followed by a sequence of sparse residual blocks. The sparse residual block contains an identity connection and a parallel residual connection, which operates on the sparse tensors. The residual connection is composed of two $3 \times 3 \times 3$ submanifold sparse convolutions \cite{submanifold} with stride 1 for learning the residual features. 

\begin{figure}[!tp] 
	\centering
    \includegraphics[height=5cm]{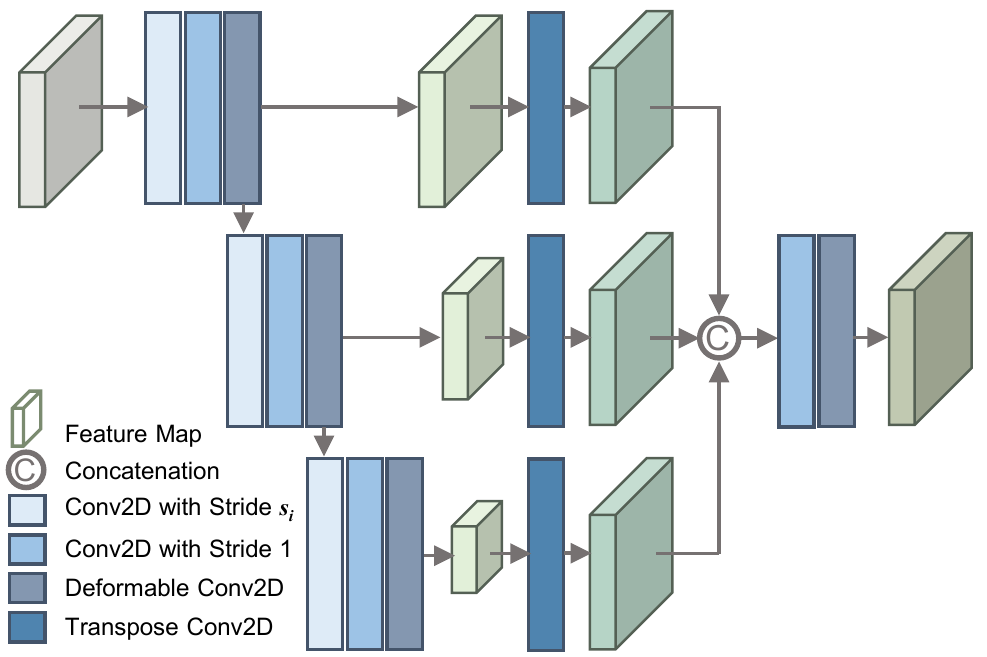}%
    \vspace{-2mm}
	\caption{Illustration of the deformable convolutional tower.}
	\vspace{-4mm}
	\label{2Dconv_tower}
\end{figure}

\subsection{Attentive Deformable Feature Adaptation}
\textbf{Feature-level 3D-to-2D transformation.}
From ResVoxelNet, we can generate 3D feature volume in a shape as $(C_{1}, L, W, H)$. It takes too much computational cost when it is processed further or directly followed by a detection head. Considering that the object still maintains its original physical size without distortion and overlaps when it is projected into the BEV, the 3D feature volume can be stacked on the height dimension to form a 2D projection in the BEV following \cite{Second, SASSD, PVRCNN}. Thus, the 3D feature volume can be reshaped as $(C_{1}*H, L, W)$. Since the features of the height dimension are merged into the channel dimension, the height information is still retained. Then, we can employ the 2D convolutional operations to handle the feature embeddings.  

\textbf{Deformable convolutional tower.}
The deformable convolutional tower aggregates the features at different scales and adaptively adjusts the receptive fields with the deformable convolution \cite{MDCN}. As shown in Fig.~\ref{2Dconv_tower}, the convolutional tower contains three levels that share the same structure. The first layer of each level applies a $3 \times 3 $ convolutional layer with stride $s_i$ to downsample the feature map, followed by several $3 \times 3 $ convolutional layers with the stride of 1. We then use a $3 \times 3$ deformable convolutional layer to adaptively adjust the receptive fields at each position on the feature map. Next, the feature maps from the three branches are upsampled to the desired resolution and then concatenated together. Before using another deformable convolutional layer to further enhance the feature learning, we employ another $3 \times 3$ convolutional layer with the stride of 1 to reduce the channels for computational efficiency. Each convolutional layer is followed by the BatchNorm and ReLU operations. We set the three-level down-sampling stride $s_i$ to 1, 2, 2, and employ three transpose convolutions with strides of 1, 2, 4 to upsample the features.

\begin{figure}[!tp] 
	\centering
    \includegraphics[height=5cm]{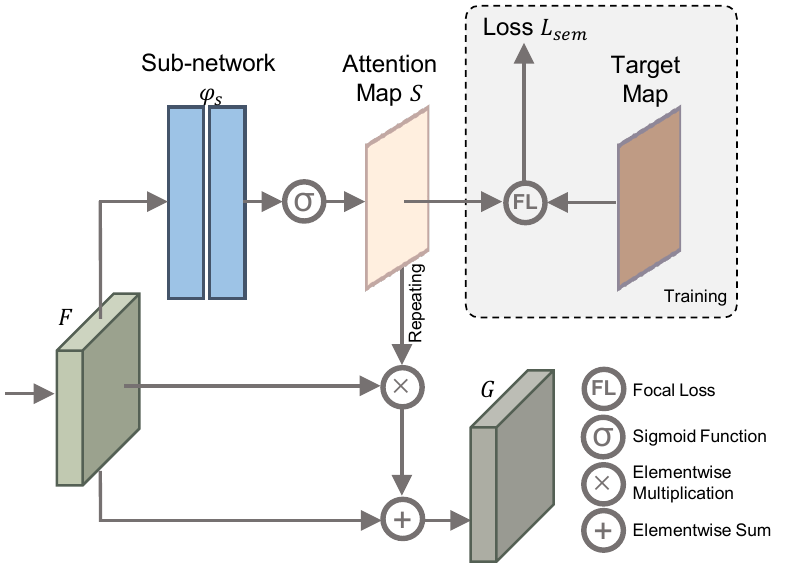}%
	\vspace{-3mm}
	\caption{The supervised mask-guided attention network.}
	\vspace{-3mm}
	\label{spatial_attention}
\end{figure}

\textbf{Learning with mask-guided attention.}
The point cloud object detection is sensitive to noise \cite{tanet_AAAI}. Since noise is inevitable in the real world, it is necessary to introduce attention to strengthen the clues from objects and weaken the clues from non-objects for robust feature learning. We introduce a supervised mask-guided attention mechanism to highlight the object pixels from the complex background. As shown in Fig.~\ref{spatial_attention}, the feature map $F$ is fed into a sub-network $ {\phi}_{s} $ with the sigmoid function $\sigma$ to predict a class agnostic segmentation mask distributed between values of 0 and 1, which can serve as the attention map $ S \in { [0, 1] }^{1 \times L \times W} $. The foreground pixels on the objects and the background pixels are expected to have a value close to 1 and 0, respectively. Hence, the objects can be highlighted from the background as 
\begin{equation}
    S = \sigma {\phi_{s}}(F), \quad \space \space G = (1 + S) F. \label{indentity_connection}
\end{equation}
An identity connection is applied in Eq. (\ref{indentity_connection}) by adding $F$, which aims to preserve the background context information.

The sub-network $\phi_{s}$ is composed of a $3\times3$ convolutional layer followed by the BatchNorm and ReLU operations for feature embedding and a $1\times1$ convolutional layer for the output. During training, a target segmentation mask can be generated from the provided ground truth boxes by checking whether the pixel is in the ground truth box in the BEV. The focal loss \cite{RetinaNet} with default settings is employed to handle the unbalanced number of the foreground and background pixels with the loss term $L_{sem}$. 

\subsection{Anchor-free Detection Head}
The proposed model can learn a dense mask-guided 2D feature map $G$ in $C_{2} \times L \times W$ dimension from the point cloud. The height information is retained in the 2D feature map by flatting the features from the height dimension into the channel dimension. Inspired by \cite{Centernet}, we construct a new single-stage anchor-free 3D detection head. The $K$-class object detection task is decomposed into three sub-tasks: $K$-class center point classification in the BEV, 3D bounding box regression based on the center points, and estimation of the IoU-based detection confidence. The third part of Fig.~\ref{overall_arch} shows three parallel heads in a structure similar to the sub-network $\phi_{s}$.

\textbf{Center point classification.} 
The center point classification head treats the objects as center points. Given the 2D feature map $G \in \mathbb{R}^{C_{2} \times L \times W}$ in the BEV, a heatmap $\hat{P} \in {\left[0,1 \right]}^{K \times L \times W}$ is predicted to represent the probability of the center point of each object. The local peaks of the $k$-th channel can be detected as the center points representing the objects of class $k$. And the center point can be directly decoded as the object's center coordinates $(x,y)$ in the BEV.

The target heatmap $P \in {\left[0,1 \right]}^{K \times L \times W}$ can be generated following \cite{Cornernet, Centernet}. For each ground truth bounding box of class $k$, the pixel index $\tilde{p} \in \mathbb{R}^2$ of the center point $p \in \mathbb{R}^{2}$ on the feature map can be calculated as $\tilde{p} = \lfloor \frac{p}{R} \rfloor$, where $R$ is the output stride. Then a 2D Gaussian exp$( -\frac{(x - \tilde{p_x})^2 + (y - \tilde{p_y})^2}{2\sigma_{p} ^2} )$ is placed around $\tilde{p}$ to form the target heatmap, where $\sigma_{p}$ is a box-size adaptive standard deviation \cite{Cornernet}. Only the pixels $\tilde{p}$ at center points are positive, and all other pixels are negative. We use a variant of focal loss to supervise the center classification task as:
\begin{equation}
    L_{\text{cls}}=\frac{-1}{N} \sum_{i}\left\{\begin{array}{ll}
    \left(1-\hat{P}_{i}\right)^{\alpha} \log \left( \hat{P}_{i}\right) & \text { if } {P}_{i}=1, \\
    \left(1-{P}_{i}\right)^{\beta} \hat{P}_{i}^{\alpha} \log \left(1-\hat{P}_{i}\right) & \text { otherwise, }
    \end{array}\right.
\end{equation}
where $i$ indexes the pixel location on the heatmap, $N$ is the number of objects, and the hyperparameters $\alpha$ and $\beta$ are selected as $\alpha=2$ and $\beta=4$ \cite{Cornernet}. Please refer to Fig.~\ref{visualization_ADFA}.(h) - (j) for a clear visualization.

\textbf{3D Bounding box regression.}
The 3D object detection locates an object as $(x, y, z, l, w, h, \theta)$, where $(x, y, z)$, $(l, w, h)$, and $\theta$ are denoted as the 3D position coordinates, 3D bounding box size, and the rotation angle of the object. 

Although the $(x, y)$ can be decoded from the integer coordinates of the detected center point from the 2D feature map, we can predict an offset $({\Delta x}, {\Delta y})$ to compensate for the difference between the actual floating-point coordinates and the integer coordinates of the center point. Since the height information is still retained in the 2D feature map, the coordinate $z$ and the size $(l,w,h)$ are directly regressed. The predictions of the offset $({\Delta x}, {\Delta y})$, the coordinate $z$ and the size $(l,w,h)$ are all trained with the $\text{L1}$ loss:
\begin{align}
    L_{\text{l1}} = \frac{1}{N} \sum_{\tilde{p}} \left| \hat{A}_{\tilde{p}} - A_{\tilde{p}} \right|,
\end{align}
where only the center point locations $\tilde{p}$ on the prediction map $\hat{A}$ and that on the target map $A$ are computed for the loss. All other locations are ignored.

To ease the regression of the rotation angle without any predefined oriented anchors, we divide $(0, 2\pi)$ into $b$ bins and use the bin-res loss \cite{PointRCNN} to conduct the supervision. The model first roughly classifies which bin the $\theta$ is in, and then finely regresses the residuals in this bin. For each center point location $\tilde{p}$, the rotation angle loss $L_{\text{rot}, \tilde{p}}$ can be composed of the bin classification loss $L^{\text{bin}}_{\text{rot}}$ and the residual regression loss $L^{\text{res}}_{\text{rot}}$ as
\begin{align}
    L_{\text{rot}, \tilde{p}} =  L^{bin}_{\text{rot}}( \hat{R}^{\text{bin}}_{\tilde{p}}, R^{\text{bin}}_{\tilde{p}} ) + L^{\text{res}}_{\text{rot}}( \hat{R}^{\text{res}}_{\tilde{p}}, R^{\text{res}}_{\tilde{p}} ), \label{rot_loss_p}
\end{align}
where $\hat{R}^{\text{bin}}$ and $\hat{R}^{\text{res}}$ are predicted bin classes and residuals while $R^{\text{bin}}$ and $R^{\text{res}}$ are their targets. The total rotation angle loss of $N$ objects can be formulated as 
\begin{align}
    L_{\text{rot}} = \frac{1}{N} \sum_{\tilde{p}} L_{\text{rot}, \tilde{p}} . 
\end{align}

Moreover, a corner loss is used to supervise all items related to the 3D bounding box as a whole. The corner loss $L_{\text{corner}}$ is defined as the $\text{L2}$ distance between the eight corners of the predicted bounding box and the assigned ground truth as
\begin{equation}
    L_{\text{corner}}= \frac{1}{N} \sum_{\tilde{p}}( \sum_{k=1}^{8}\left\|\hat{B}_{\tilde{p},k} - B_{\tilde{p},k}\right\| ),
\end{equation}
where $\tilde{p}$ indicates that the $L_{\text{corner}}$ only considers the bounding boxes regressed from the center point locations. Therefore, the bounding box prediction is supervised by 
\begin{align}
    L_{\text{box}} = \gamma_{1} L_{\text{offset}} + \gamma_{2} L_{\text{z-coord}} + \gamma_{3} L_{\text{size}} + \gamma_{4} L_{\text{rot}} + \gamma_{5} L_{\text{corner}}, 
\end{align}
where $\gamma_{1} \sim \gamma_{5}$ are the constant coefficients to balance each loss.

\textbf{IoU-based detection confidence recalibration.}
To detect an object with a confidence score, we can directly use the local peak value on the center point classification heatmap as detection confidence of the object \cite{Centernet}. The model can achieve better detection performance when the detection confidence is positively correlated with the IoU of the predicted bounding box \cite{IoUNet,GFocal}. However, the local peak values on the classification heatmap are all close to 1 and have a low positive correlation with the actual IoUs of the predicted bounding boxes. To improve the correlation, a parallel confidence estimation head is designed to predict IoU-based confidence. The detection confidence is re-calibrated from the classification score to the IoU-based confidence.

The training samples and targets for the confidence estimation head are obtained automatically during the training of the network. The top $M$ local peaks on the predicted classification heatmap are utilized to select the corresponding $M$ predicted bounding boxes as the confidence training samples. $M$ is a hyperparameter we chose to balance the training samples with IoUs greater than zero and equal to zero. For the $m$-th bounding box, we calculate the 3D IoUs between the $m$-th bounding box and all ground truth bounding boxes, and then assign the largest one as its actual IoU denoted as $IoU_{m}$. We follow \cite{PVRCNN} to normalize each confidence training target $c_m$ between $\left[ 0, 1 \right]$ as:
\begin{align}
    c_m = \text{min}(1, \text{max}(0, 2IoU_{m} -0.5)).
\end{align}
The confidence estimation head is then trained to minimize the cross-entropy loss computed from the predicted confidence $\hat{c}_{m}$ and its target $c_m$ as:
\begin{align}
    L_{\text{iou}} = \frac{-1}{M}\sum_{m} ( c_{m} \text{log}(\hat{c}_{m}) + (1-c_m) \text{log}(1 - \hat{c}_{m} ) ).
\end{align}
In the inference stage, the confidence $\hat{c}$ predicted by the network can be directly assigned to the corresponding predicted bounding box as the detection confidence score. 

The two-stage methods PV-RCNN \cite{PVRCNN} and STD \cite{STD} also learn the IoUs of the proposal boxes to guide the NMS. Note that there are two main differences in our method: 1) PV-RCNN and STD require the actual IoUs of the proposal boxes as the training target for their confidence branches in the second stage. However, there are no proposal boxes or anchor boxes in our one-stage anchor-free methods. Instead, we use IoUs of the predicted boxes as training targets for IoU prediction in a parallel regression head. 2) PV-RCNN and STD use the predicted IoUs for refined boxes de-redundancy in NMS, while we only re-calibrate the detection confidence using the predicted IoUs without NMS.

\textbf{Training loss.}
Our network is trained end-to-end with a multi-task loss $L_{\text{total}}$ as Eq.(\ref{loss_total}), which is equally contributed by the center point classification loss $L_{\text{cls}}$, the 3D bounding box regression loss $L_{\text{box}}$, the IoU-based confidence training loss $L_{\text{iou}}$, and the mask-guided attention training loss $L_{\text{sem}}$. 
\begin{align}
    L_{\text{total}} = L_{\text{cls}} + L_{\text{box}} + L_{\text{iou}} + L_{\text{sem}} \label{loss_total} .
\end{align}

\section{Experiment}
\subsection{Datasets}
\textbf{KITTI Dataset.}
KITTI 3D object detection benchmark \cite{KITTIleaderboard} is widely used for 3D object detection evaluation.
The KITTI dataset \cite{KITTIDataset} provides 7481 training samples and 7518 testing samples with objects of three classes: car, pedestrian, and cyclist. The training samples are generally divided into the \textit{train} split set (3712 samples) and the \textit{val} split set (3769 samples) for experiments \cite{MV3D, VoxelNet, PointRCNN}. For each class, the detection performance is evaluated as the Average Precision (AP) based on three difficulty levels: easy, moderate, and hard. Following the official evaluation protocol, we calculate the $\text{AP}_{\text{3D}}$ with 40 recall positions as the evaluation metric.

\textbf{Waymo Open Dataset.} The newly released Waymo Open Dataset \cite{waymo_open_dataset} is the currently largest public dataset for autonomous driving. Waymo contains $\sim$158K point cloud training samples and $\sim$40K point cloud validation samples. The objects in Waymo are divided into two levels: LEVEL\_1 for boxes with more than five LiDAR points, and LEVEL\_2 for boxes with at least one LiDAR point. The official evaluation tool provides metrics of $\text{AP}_{\text{3D}}$ and $\text{AP}_{\text{BEV}}$ for the 3D and BEV detection.

\subsection{Implement Details}
\textbf{Network architecture.}
Since the objects in KITTI are only annotated in the camera Field of Vision (FOV), the quantization step $d$ is set as $(0.05, 0.05, 0.1)$ meters to voxelize the KITTI point clouds within the range of $\left[ 0, 70.4 \right] \times \left[ -40, 40 \right] \times \left[ -3, 1 \right]$ meters along the $X$, $Y$, $Z$ axis, resulting in an initial spatial resolution $(L_{0}, W_{0}, H_{0})$ of $(1408, 1600, 40)$. For the full \ang{360}-field detection in Waymo, we voxelize the Waymo point clouds within the range of $\left[ -75.2, 75.2 \right] \times \left[ -75.2, 75.2 \right] \times \left[ -2, 4 \right]$ meters with the quantization step $d$ as $(0.1, 0.1, 0.15)$ along the $X$, $Y$, $Z$ axis, resulting in an initial spatial resolution $(L_{0}, W_{0}, H_{0})$ of $(1504, 1504, 40)$. The ResVoxelNet downsamples the feature volumes with $1 \times$, $2 \times $ $4 \times$, $8 \times$ while increases the feature dimension as 16, 32, 64, 128. The ADFA module maintains the $8 \times$ downsampled resolution $(L, W)$ of (176, 200) or (188, 188) with a feature dimension $C_{2}$ of 256 for KITTI and Waymo models. Each head is implemented by two convolutions with stride of 1: a $3 \times 3$ convolution with 128 channels, and a $ 1\times 1$ convolution with different numbers of channels for different output values.

\textbf{Training details.} 
Our model is trained from scratch in an end-to-end manner with the AdamW \cite{AdamW} optimizer and one-cycle policy \cite{one_cycle_lr} with LR max 0.01, division factor 10, momentum ranges from 0.95 to 0.85, weight decay 0.01. For KITTI, the model is trained for 50 epochs with batch size 8 on a Tesla V100 GPU. For Waymo, the model is trained for 36 epochs with batch size 36 on 4 Tesla V100 GPUs. To avoid overfitting during the training process, four commonly used data augmentation strategies are used to avoid overfitting: ground truth sampling \cite{Second}, random flipping along the $X$ axis, global scaling with a random scaling factor in $[0.95, 1.05]$, global rotation around the $Z$ axis with a random angle in $[- \frac{\pi}{4}, \frac{\pi}{4} ]$. 

\textbf{Inference details.} 
For the inference, the Non-maximum suppression (NMS) post-processing process is no longer needed \cite{Centernet}. A $3 \times 3$ max-pooling operation is applied to extract the local peaks on each channel of the center point classification heatmap. The detection results at the top 50 or 200 local peaks are kept as candidates for KITTI and Waymo since the Waymo point cloud contains more object instances than KITTI. A threshold $\mu_{cls}$ of 0.5 is then applied to filter the negative detection results.

\begin{figure*}[t!]
	\centering
	\includegraphics[width=0.9\linewidth]{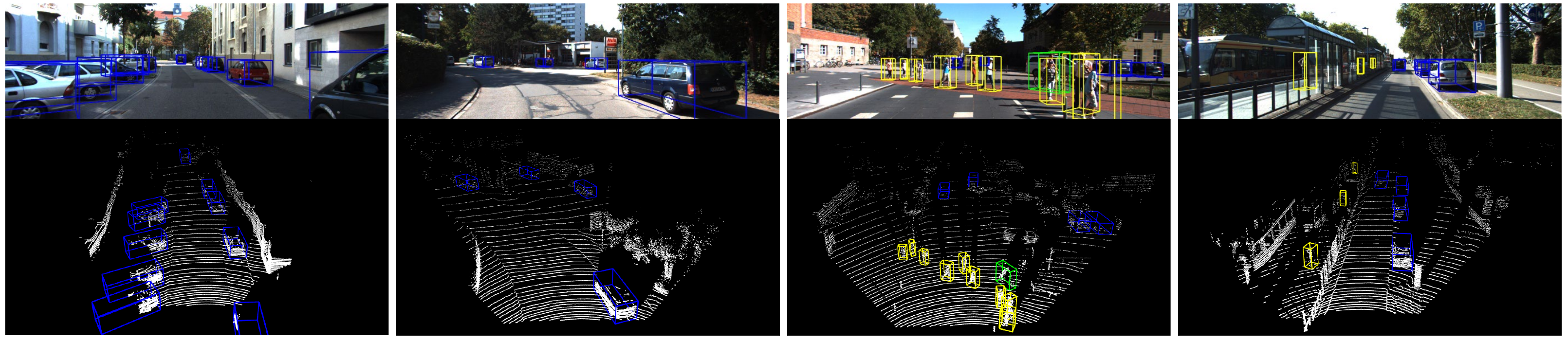}%
	\vspace{-2mm}
	\caption{Qualitative results of our method on the KITTI \textit{test} set. We show the predicted 3D bounding boxes of cars (blue), cyclists (green) and pedestrians (yellow) on both the image and the point cloud. Best viewed with color and zooming in.}
	\label{qualitative_results}
\end{figure*}

\begin{table}[!tbp]
\caption{Performance comparison on car class of the KITTI \textit{test} set. Methods are divided into two types: two-stage and single-stage. Blue numbers are results for the best detection performance. Bold numbers denote the top-2 results for single-stage point cloud detectors.}
\vspace{-3mm}
\label{AP_car_test}
\begin{center}
\setlength{\tabcolsep}{0.44mm}
\begin{tabular}{c|c|c|ccc}
\toprule
\multirow{2}{*}{Method} & \multirow{2}{*}{Sensor}  & \multirow{2}{*}{Anchor} & \multicolumn{3}{c}{$\text{AP}_{\text{3D}}$}  \\
                        &                          &                              & Easy    & Mod.   & Hard   \\
\midrule                        
MV3D\cite{MV3D}                     & LiDAR + RGB   & $\surd$  & 74.97   & 63.63  & 54.00  \\
F-PointNet\cite{F-PointNets}        & LiDAR + RGB   & $\surd$  & 82.19   & 69.79  & 60.59  \\
PointPainting\cite{PointPainting}   & LiDAR + RGB   & $\surd$  & 82.11   & 71.70  & 67.08  \\
AVOD\cite{AVOD}                     & LiDAR + RGB   & $\surd$  & 83.07   & 71.76  & 65.73  \\
PI-RCNN\cite{PIRCNN}                & LiDAR + RGB   & $\surd$  & 84.37   & 74.82  & 70.03  \\
CLOCs\_PointCas\cite{CLOCs}          & LiDAR + RGB   & $\surd$  & 87.50   & 76.68  & 71.20  \\
MMF\cite{MMF}                       & LiDAR + RGB   & $\surd$  & 88.40   & 77.43  & 70.22  \\
PointRCNN\cite{PointRCNN}           & LiDAR only    & $\surd$  & 86.96   & 75.64  & 70.70 \\
Fast PointRCNN\cite{Fast-PointRCNN} & LiDAR only    & $\surd$  & 85.29   & 77.40  & 70.24 \\
STD\cite{STD}                       & LiDAR only    & $\surd$  & 87.95   & 79.71  & 75.09 \\
Part-A2\cite{PartA2_TPAMI}          & LiDAR only    & $\surd$  & 87.81   & 78.49  & 73.51 \\
PV-RCNN\cite{PVRCNN}                & LiDAR only    & $\surd$  &\textcolor{blue}{90.25}    &\textcolor{blue}{81.43}  &\textcolor{blue}{76.82}  \\
\midrule
\midrule
VoxelNet\cite{VoxelNet}             & LiDAR only    & $\surd$  & 78.82   & 64.17  & 57.51 \\ 
SECOND\cite{Second}                 & LiDAR only    & $\surd$  & 83.34   & 72.55  & 65.82 \\ 
PointPillars\cite{PointPillars}     & LiDAR only    & $\surd$  & 82.58   & 74.31  & 68.99 \\
SCNet\cite{SCNet}                   & LiDAR only    & $\surd$  & 83.34   & 73.17  & 67.93 \\
TANet\cite{tanet_AAAI}              & LiDAR only    & $\surd$  & 84.39   & 75.94  & 68.82 \\
Associate-3Ddet\cite{Associate-3Ddet}& LiDAR only   & $\surd$  & 85.99   & 77.40  & 70.53 \\
Point-GNN\cite{PointGNN}            & LiDAR only    & $\surd$  & \textbf{88.33}   & \textbf{79.47}  & 72.29 \\
HotSpotNet\cite{Hotspots(ECCV2020)} & LiDAR only    & $\times$   & 87.60   & 78.31  & \textbf{73.34} \\
\textbf{Ours}                       & LiDAR only    & $\times$   & \textbf{88.16}   & \textbf{79.68}  & \textbf{72.39}  \\
\bottomrule
\end{tabular}
\end{center}
\vspace{-2mm}
\end{table}

\begin{table}[!tbp]
\caption{$\text{AP}_{\text{3D}}$ comparison on cyclist and pedestrian classes of the KITTI \textit{test} set. Methods are divided into two types: two-stage and single-stage. Blue numbers are results for the best detection performance. Bold numbers denote the top-2 results for single-stage point cloud detectors.}
\vspace{-3mm}
\label{AP_cyc_ped_test}
\small
\begin{center}
\resizebox{0.48\textwidth}{!}{
\begin{tabular}{c|c|ccc|ccc}
\toprule
\multirow{2}{*}{Method} &\multirow{2}{*}{Anchor} & \multicolumn{3}{c|}{Cyclist}  & \multicolumn{3}{c}{Pedestrian}\\
                        &                        & Easy    & Mod.   & Hard      & Easy    & Mod.   & Hard\\
\midrule
AVOD-FPN\cite{AVOD}             & $\surd$   & 63.76   & 50.55  & 44.93  & 50.46  & 42.27   & 39.04 \\ 
F-PointNet\cite{F-PointNets}    & $\surd$   & 72.27   & 56.12  & 49.01  & 50.53  & 42.15   & 38.08 \\ 
STD\cite{STD}                   & $\surd$   & 78.69   & 61.59  & 55.30   &53.29 &42.47 &38.35 \\
PointPainting\cite{PointPainting} & $\surd$ & 77.63   & \textcolor{blue}{63.78}  & 55.89   & 50.32  & 40.97  & 37.87 \\ 
PV-RCNN\cite{PVRCNN}            & $\surd$   & 78.60   & 63.71  & \textcolor{blue}{57.65}  & N/A  & N/A  & N/A\\ 
\midrule
\midrule
SCNet\cite{SCNet}               & $\surd$   & 67.98   & 50.79  & 45.15  & 47.83  & 38.66  & 35.70 \\ 
PointPillars\cite{PointPillars} & $\surd$   & \textbf{77.10}   & 58.65  & 51.92  & \textbf{51.45} & 41.92 & 38.89 \\ 
SemanticVoxels\cite{semanticvoxels} & $\surd$ & N/A     & N/A   & N/A   & 50.90 & 42.19 & 39.52 \\ 
TANet\cite{tanet_AAAI}          & $\surd$   & 75.70   & \textbf{59.44}  & \textbf{52.53}   & \textcolor{blue}{\textbf{53.72}}  & \textcolor{blue}{\textbf{44.34}}  & \textcolor{blue}{\textbf{40.49}}\\ 
\textbf{Ours}                   & $\times$   &\textcolor{blue}{\textbf{80.64}}   & \textbf{63.43}  & \textbf{55.15}  & 50.65  & \textbf{43.09} & \textbf{39.65}\\ 
\bottomrule
\end{tabular}
}
\end{center}
\vspace{-2mm}
\end{table}

\begin{figure}[!tp] 
	\centering
    \includegraphics[height=4.0cm]{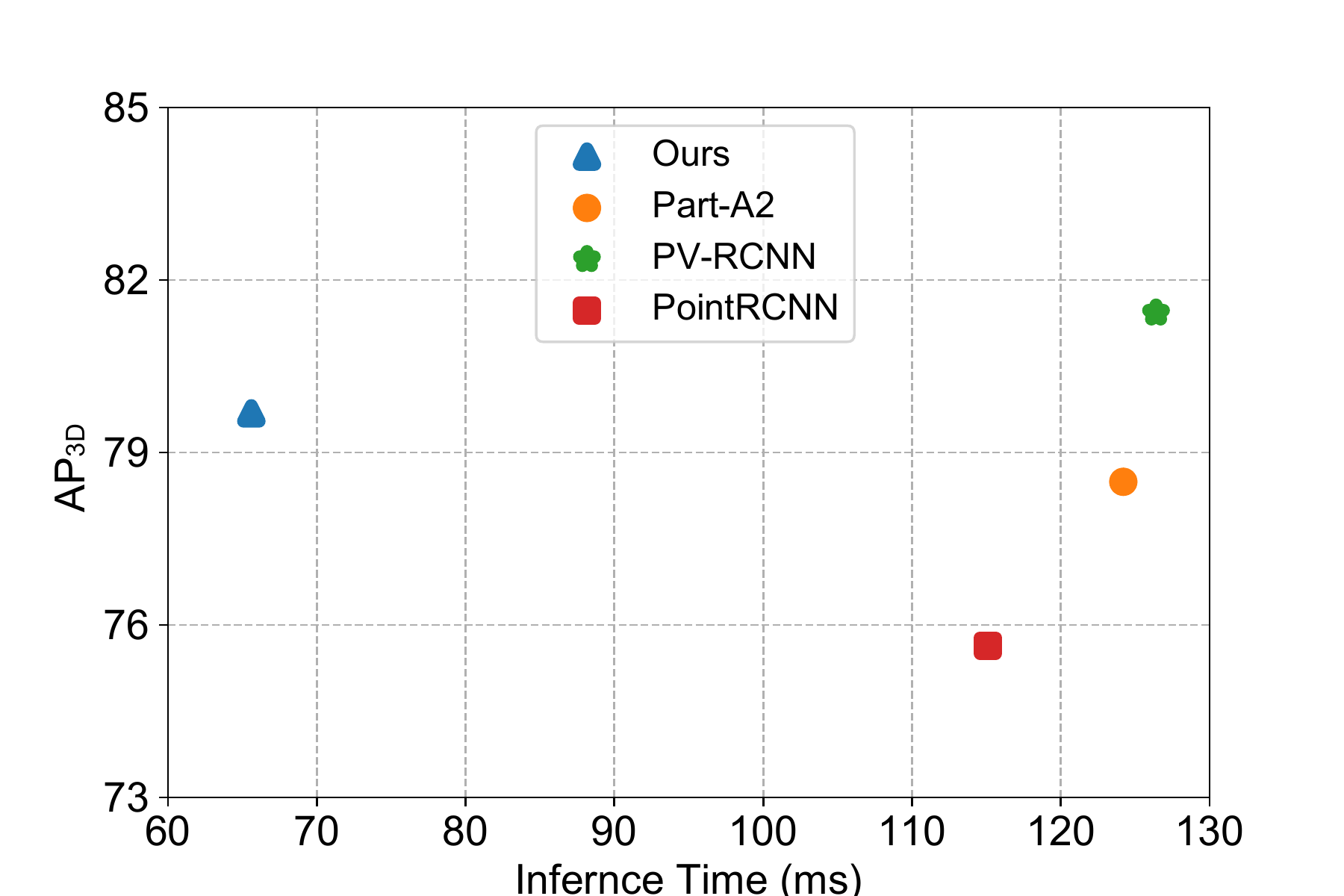}%
	\vspace{-2mm}
	\caption{Latency comparison: $\text{AP}_{\text{3D}}$ on moderate car class of KITTI \textit{test} set (top consideration in KITTI) vs. inference time in milliseconds.}
	\vspace{-3mm}
	\label{ap_time_scatter}
\end{figure}

\subsection{Results on KITTI Dataset}
Fig.~\ref{qualitative_results} shows the qualitative results of our method on the KITTI \textit{test} set. Tab.~\ref{AP_car_test} and Tab.~\ref{AP_cyc_ped_test} present the performance comparison between our anchor-free single-stage method and the most existing state-of-the-art methods on the KITTI \textit{test} set. For the main considered car class shown in Tab.~\ref{AP_car_test}, our method makes $\text{AP}_{\text{3D}}$ margins of 0.56\%, 1.37\% on the easy and moderate levels compared with the latest anchor-free method HotSpotNet \cite{Hotspots(ECCV2020)}, which shows that the object center is more robust than the various hotspots lying on the object surface. For $\text{AP}_{\text{3D}}$ on the three levels, our method outperforms the state-of-the-art anchor-based single-stage point cloud detectors. Our method still outperforms the most two-stage methods, and achieves better performance with the STD \cite{STD} and Part-A2 \cite{PartA2_TPAMI} on the easy level. Our method also has significant improvement when compared to MV3D \cite{MV3D} and AVOD \cite{AVOD}, indicating that the feature level 3D-to-2D transformation effectively avoids the information loss caused by the 3D-to-2D data conversion at the input level. 

In addition, our method performs well on small objects like cyclists and pedestrians, as shown in Tab.~\ref{AP_cyc_ped_test}. Most methods do not provide results for cyclists and pedestrians, due to the difficulty of detecting small objects. For cyclists, our method not only outperforms the single-stage detectors, but also achieves better performance against the two-stage anchor-based PV-RCNN \cite{PVRCNN} on the easy level. The results show the effectiveness of our method. 

\textbf{Latency comparison.}
Fig.~\ref{ap_time_scatter} demonstrates the latency comparison that is performed on the same machine with the same settings. Since most methods do not report the inference time, we have re-implemented the methods with official codes publicly available. Our method saves almost half of the inference time, while it achieves comparable performance with PV-RCNN \cite{PVRCNN} and it is even better than other two-stage methods Part-A2 \cite{PartA2_TPAMI} and PointRCNN \cite{PointRCNN}.

\subsection{Results on Waymo Open Dataset}
We evaluate the performance on the Waymo Open Dataset for both LEVEL\_1 and LEVEL\_2, and compare the $\text{AP}_{\text{3D}}$ and $\text{AP}_{\text{BEV}}$ with the top-performing methods. Tab.~\ref{ap_waymo_val} shows that we achieve better performance than the existing methods with 71.98\% $\text{AP}_{\text{3D}}$ and 87.15\% $\text{AP}_{\text{BEV}}$ on the most commonly used LEVEL\_1. Besides, our method outperforms PV-RCNN on the BEV of LEVEL\_2. The improvements on the $\text{AP}_{\text{BEV}}$ demonstrate the effectiveness of our method for detecting the objects as the center points in the BEV.

\begin{table}[!t]
\caption{Performance comparison on the Waymo validation for vehicle detection. The top-1 results are in bold.}
\label{ap_waymo_val}
\vspace{-3mm}
\small
\setlength{\tabcolsep}{0.6mm}
\resizebox{0.45\textwidth}{!}{
\begin{tabular}{c|c|c|cccc}
\toprule
Difficulty                 & Method       &Stage  & Overall & \footnotesize{{[}0, 30)m} & \footnotesize{{[}30, 50)m} & \footnotesize{ {[}50, Inf)m }\\
\midrule
\multirow{14}{*}{LEVEL\_1} &              &             & \multicolumn{4}{c}{$\text{AP}_{\text{3D}}$}                          \\
\cline{2-7}
                           & PV-RCNN\cite{PVRCNN}      &2  & 70.30    & \textbf{91.92}      & 69.21       & 42.17        \\
                           & PointPillars\cite{PointPillars} &1  & 56.62   & 81.01      & 51.75       & 27.94        \\
                           & MVF\cite{MVF}          &1  & 62.93   & 86.30       & 60.02       & 36.02        \\ 
                           & Pillar-OD\cite{pillar-od}    &1  & 69.80    & 88.53      & 66.50        & 42.93        \\ 
                           & \textbf{Ours}         &1  & \textbf{71.98}   & 91.85      & \textbf{70.58}       & \textbf{48.04}        \\
\cline{2-7}
                           &              &             & \multicolumn{4}{c}{$\text{AP}_{\text{BEV}}$}                        \\
\cline{2-7}
                           & PV-RCNN\cite{PVRCNN}    &2   & 82.96   & 97.35      & 82.99       & 64.97        \\
                           & PointPillars\cite{PointPillars} &1  & 75.57   & 92.10       & 74.06       & 55.47        \\
                           & MVF\cite{MVF}          &1  & 80.40    & 93.59      & 79.21       & 63.09        \\
                           & Pillar-OD\cite{pillar-od} &1    & 87.11   & 95.78      & 84.74       & 72.12        \\
                           & \textbf{Ours}        &1 & \textbf{87.15}       & \textbf{97.42}          & \textbf{86.80}          & \textbf{73.92}            \\
\midrule
\multirow{6}{*}{LEVEL\_2}  &              &             & \multicolumn{4}{c}{$\text{AP}_{\text{3D}}$}                          \\
\cline{2-7}
                           & PV-RCNN\cite{PVRCNN}     &2 & \textbf{65.36}   & \textbf{91.58}      & \textbf{65.13}       & \textbf{36.46}        \\
                           & \textbf{Ours}         &1  & 64.78   & 88.85      & 64.64       & 36.17        \\
\cline{2-7}
                           &              &             & \multicolumn{4}{c}{$\text{AP}_{\text{BEV}}$}                         \\
\cline{2-7}
                           & PV-RCNN\cite{PVRCNN}     &2  & 77.45   & \textbf{94.64}      & 80.39       & 55.39        \\
                           & \textbf{Ours}        &1  & \textbf{77.85}       & 94.39          & \textbf{80.87}           & \textbf{59.16}          \\
\bottomrule
\end{tabular}
}
\vspace{-2mm}
\end{table}

\subsection{Ablation Study}
Extensive ablation experiments are conducted to analyze each individual component of our methods. Following the existing works \cite{PVRCNN, PointGNN, Second, STD, PointRCNN, PIRCNN}, we train the models on the KITTI \textit{trian} split set with the same training settings and adopt the $\text{AP}_\text{3D}$ on the car class of KITTI \textit{val} split set as the performance evaluation metric. Besides, we also perform each ablation experiment on the Waymo vehicle detection and report the $\text{AP}_\text{3D}$ on the Waymo validation for stability. In each table, the best performance is marked in bold.

\begin{figure}[!t]
	\centering
	\includegraphics[width=0.95\linewidth]{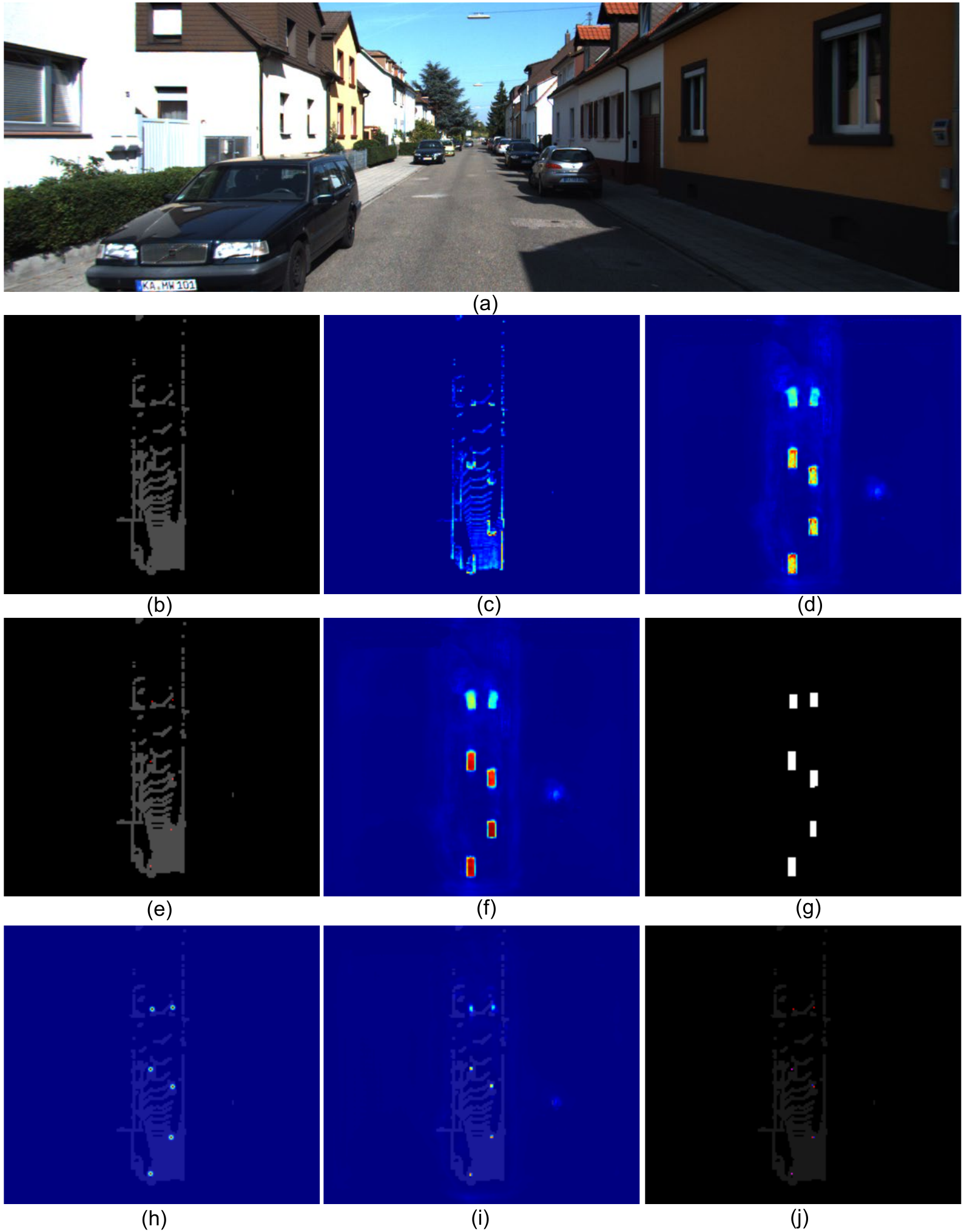}%
	\caption{Visualization of the ADFA module and center point classification. (a) RGB image. (b) Occupancy grids of the sparse 3D feature volume in the BEV. (c) Input feature map of ADFA module. (d) Output feature map of ADFA module. (e) Occupancy grids with red points of object centers. (f) Predicted mask-guided attention map. (g) Binary segmentation mask. (h) Target and (i) predicted heatmap of center point classification. (j) The local peaks from the predicted heatmap and the target heatmap are colored in blue and red, and become purple if they overlap.}
	\label{visualization_ADFA}
\end{figure}

\textbf{ADFA module.}
Our ADFA module consists of two main components: the mask-guided attention and the deformable convolutional tower that can be assembled with or without the deformable convolutions. As shown in Tab.~\ref{ADFA_module_AP}, the performance drops significantly when we remove the entire ADFA module, indicating the importance of the ADFA module. When adding a vanilla convolutional tower combined with multiple receptive fields, the performance becomes better, especially for the moderate and hard levels on KITTI. After the receptive fields are adaptively adjusted by the deformable convolutions and the feature maps are enhanced by the mask-guided attention, the performance becomes better on both KITTI and Waymo. Compared with the input feature map in Fig.~\ref{visualization_ADFA}.(c), the output feature map in Fig.~\ref{visualization_ADFA}.(d) becomes much denser and more focused on the entire object. With the dense mask-guided feature map provided by the ADFA module, our model achieves the best performance as shown in the last row of Tab.~\ref{ADFA_module_AP}.

\begin{table}[!tbp]
\caption{Effectiveness ($\text{AP}_{\text{3D}}$) of each individual component in the ADFA module. ``Conv.'', ``DCN'', and ``Att.'' denote  2D convolutional tower, deformable convolution, and mask-guided attention, respectively.}
\label{ADFA_module_AP}
\vspace{-3mm}
\begin{center}
\setlength{\tabcolsep}{1.2mm}
\resizebox{0.45\textwidth}{!}{
\begin{tabular}{ccc|ccc|cc}
\toprule
\multirow{2}{*}{Conv.} & \multirow{2}{*}{DCN} & \multirow{2}{*}{Att.} & \multicolumn{3}{c|}{KITTI} & \multicolumn{2}{c}{Waymo}\\
                            &                      &                         & Easy    & Mod.   & Hard      & LEVEL\_1  & LEVEL\_2 \\
\midrule
$\times$    &$\times$ &$\times$     & 89.87   & 78.55  & 75.92  & 70.29    & 61.20\\
$\surd$     &$\times$ &$\times$     & 89.81   & 80.97  & 78.23  & 71.17    & 62.13\\
$\surd$     &$\surd$  &$\times$     & 90.27   & 81.14  & 78.40  & 71.42  & 64.24\\
$\surd$     &$\surd$  &$\surd$      & \textbf{91.86}   & \textbf{83.29}  & \textbf{78.51}  & \textbf{71.98} & \textbf{64.78} \\
\bottomrule
\end{tabular}
}
\end{center}
\vspace{-2mm}
\end{table}

\begin{table}[!tbp] 
\caption{Effectiveness of supervised mask-guided attention.}
\vspace{-3mm}
\label{semantic_attention_AP}
\begin{center}
\setlength{\tabcolsep}{0.8mm}
\resizebox{0.45\textwidth}{!}{
\begin{tabular}{cc|ccc|cc}
\toprule
\multirow{2}{*}{\small{Supervision}} & \multirow{2}{*}{ \small{Reweighting}} & \multicolumn{3}{c|}{KITTI} & \multicolumn{2}{c}{Waymo}\\
                       &                           & Easy    & Mod.   & Hard   & LEVEL\_1  & LEVEL\_2 \\
\midrule
$\times$               &$\times$                   & 90.27   & 81.14  & 78.40  & 71.42  & 64.24\\
$\surd$                &$\times$                   & 90.52   & 83.09  & 78.22  & 71.57  & 64.39\\
$\surd$                &$\surd$                    & \textbf{91.86}   & \textbf{83.29}  & \textbf{78.51}  & \textbf{71.98} & \textbf{64.78}\\
\bottomrule
\end{tabular}
}
\end{center}
\vspace{-2mm}
\end{table}

\textbf{Supervised mask-guided attention.}
Tab.~\ref{semantic_attention_AP} validates the effectiveness of supervised mask-guided attention. In multi-task learning, the related auxiliary tasks usually facilitate the learning of the main task. As shown in the second row of Tab.~\ref{semantic_attention_AP}, the mask-guided supervision contributes to the improvement on KITTI and Waymo. When we further use the mask-guided attention map to penalize the feature map, our method achieves much higher $\text{AP}_{\text{3D}}$. Fig.~\ref{visualization_ADFA}.(f) illustrates the predicted mask-guided attention map.

\textbf{Detection confidence recalibration.}
We investigate the importance of the IoU-based detection confidence recalibration in three models with different settings shown in Tab.~\ref{confidence_recalib_AP} and Tab.~\ref{confidence_recalib_AP_waymo}. From model A to model B, the joint learning of the auxiliary IoU-based confidence estimation task brings the slight improvements of 0.72\%, 0.61\%, 0.45\% $\text{AP}_{\text{3D}}$ on KITTI and 0.56\%, 0.53\% $\text{AP}_{\text{3D}}$ on Waymo. The comparison between the model B and C with the same trained network demonstrates that the proposed IoU-based detection confidence recalibration scheme significantly improves the model B with 2.90\%, 2.17\%, 1.50\% $\text{AP}_{\text{3D}}$ on KITTI and 1.94\%, 1.71\% $\text{AP}_{\text{3D}}$ on Waymo, respectively. Moreover, we calculate the correlation between the confidence of the detection results and their actual 3D IoUs to analyze how the confidence recalibration mechanism works. The highest Pearson Linear Correlation Coefficient (PLCC) and Spearman Rank-order Correlation Coefficient (SRCC) in the third rows of Tab.~\ref{confidence_recalib_AP} and Tab.~\ref{confidence_recalib_AP_waymo} indicates that the recalibrated confidence becomes more relevant to the detection accuracy. 

\begin{table}[!tbp]
\caption{Effectiveness of detection confidence recalibration on KITTI. ``Head$_{\text{iou}}$'' and ``Recalib.'' denote IoU-based detection confidence estimation head and recalibration.}
\vspace{-3mm}
\label{confidence_recalib_AP}
\begin{center}
\setlength{\tabcolsep}{0.7mm}
\begin{tabular}{c|cc|cc|ccc}
\toprule
\multirow{2}{*}{ }  &\multicolumn{2}{c|}{Setting}  & \multirow{2}{*}{PLCC} & \multirow{2}{*}{SRCC} & \multicolumn{3}{c}{KITTI}\\
                        & Head$_{\text{iou}}$ & Recalib.  &                       &                       & Easy  & Mod.   & Hard  \\
\midrule
A &$\times$    &$\times$   & 0.478         & 0.552         & 88.24 & 80.51 & 76.56 \\
B &$\surd$     &$\times$   & 0.485         & 0.556         & 88.96 & 81.12 & 77.01 \\
C &$\surd$     &$\surd$    & \textbf{0.547}         & \textbf{0.649}         & \textbf{91.86}  & \textbf{83.29} & \textbf{78.51}  \\
\bottomrule
\end{tabular}
\end{center}
\vspace{-2mm}
\end{table}

\begin{table}[!tbp]
\caption{Effectiveness of detection confidence recalibration on Waymo.}
\vspace{-3mm}
\label{confidence_recalib_AP_waymo}
\begin{center}
\setlength{\tabcolsep}{0.7mm}
\begin{tabular}{c|cc|cc|cc}
\toprule
\multirow{2}{*}{ }  &\multicolumn{2}{c|}{Setting}  & \multirow{2}{*}{PLCC} & \multirow{2}{*}{SRCC} & \multicolumn{2}{c}{Waymo}\\
                        & Head$_{iou}$ & Recalib.  &                       &                       & LEVEL\_1  &  LEVEL\_2 \\
\midrule
A &$\times$    &$\times$   & 0.337         & 0.339         & 69.48 & 62.54  \\
B &$\surd$     &$\times$   & 0.340         & 0.342         & 70.04 & 63.07  \\
C &$\surd$     &$\surd$    & \textbf{0.579}         & \textbf{0.689}         & \textbf{71.98} & \textbf{64.78}  \\
\bottomrule
\end{tabular}
\end{center}
\vspace{-2mm}
\end{table}

\makeatletter
\newcommand\figcaption{\def\@captype{figure}\caption}
\newcommand\tabcaption{\def\@captype{table}\caption}
\makeatother

\begin{figure}[tb]
		\begin{minipage}[h]{0.48\linewidth} 
			\tabcaption{Detection performance $\text{AP}_{\text{3D}}$ with different number ($M$) of training samples of $\text{Head}_{\text{iou}}$ on KITTI.}
            \vspace{2mm}
			\label{hyperparameterM}
			\centering
			\resizebox{1.0\textwidth}{!}{
                \begin{tabular}{c|ccc|c}
                \toprule
                \multirow{2}{*}{M} & \multicolumn{3}{c|}{KITTI}                        & \multirow{2}{*}{Mean} \\
                                   & Easy           & Mod.           & Hard           &                       \\
                \midrule
                4                & 90.22          & 80.74          & 78.10           & 83.02              \\
                8                & 90.33          & 83.10          & \textbf{78.56} & 83.99              \\
                16               & \textbf{92.03} & 83.16          & 78.42          & 84.53              \\
                24               & 91.86          & \textbf{83.29}          & 78.51          & \textbf{84.55}     \\
                32               & 90.35          & 83.24          & 78.45          & 84.01              \\
                40               & 90.37          & 83.22          & 78.36          & 83.98              \\
                48               & 90.39          & 81.42          & 78.50          & 83.43              \\
                \bottomrule
                \end{tabular}
            }
            \vspace{-2mm}

		\end{minipage}%
	\hfill
		\begin{minipage}{0.48\linewidth}
			\tabcaption{Detection performance $\text{AP}_{\text{3D}}$ with different number ($M$) of training samples of $\text{Head}_{\text{iou}}$ on Waymo.}
            \vspace{2mm}
			\label{hyperparameterM_waymo}
			\centering
			\resizebox{1.0\textwidth}{!}{
                \begin{tabular}{c|cc|c}
                \toprule
                \multirow{2}{*}{M} & \multicolumn{2}{c|}{Waymo}       & \multirow{2}{*}{Mean} \\
                                   & LEVEL\_1       & LEVEL\_2       &                       \\
                \midrule
                32                 & 71.39          & 62.32          & 66.35                 \\
                56                 & 71.54          & 64.36          & 67.95                 \\
                80                 & 71.90          & 64.70          & 68.30                 \\
                104                & \textbf{71.98} & \textbf{64.78} & \textbf{68.38}        \\
                128                & 71.94          & 64.74          & 68.34                 \\
                152                & 71.67          & 64.48          & 68.09                \\
                176                & 71.12          & 62.07          & 66.60                \\
                \bottomrule
                \end{tabular}
            }
            \vspace{-2mm}

		\end{minipage}

\end{figure}

\textbf{Hyperparameter $M$.}
As shown in Tab.~\ref{hyperparameterM}, the performance is relatively stable with the $M$ around 24 on KITTI but drops as the $M$ becomes larger or smaller. Generally, each scene of the KITTI dataset does not contain many object instances. A large $M$ may include too many negative detection results with IoUs of zero, which causes that the IoU training loss on the positive detection results is overwhelmed by that on the negative detection results. A small $M$ may include many training samples with positive IoUs, which is not conducive to the IoU-based confidence estimation head for distinguishing the false positives with IoUs of zero during the inference. According to the highest mean $\text{AP}_\text{3D}$ shown in Tab.~\ref{hyperparameterM} and Tab.~\ref{hyperparameterM_waymo}, we choose $M$ as 24 and 104 for KITTI and Waymo, respectively. The reason why the $M$ on Waymo is larger than KITTI is that the full \ang{360}-field scene in Waymo contains more object instances than the camera FOV in KITTI.

\section{Conclusion}
In this work, we propose a novel anchor-free single-stage 3D detection framework. The proposed ADFA module enables the center of the object to better perceive the whole object with the adaptive receptive fields and the supervised mask-guided attention. We design an anchor-free 3D detection head to get rid of hand-crafted anchors and NMS for box de-redundancy without sacrificing the detection accuracy. We propose an IoU-based detection confidence re-calibration scheme to improve the correlation between the detection confidence and the detection accuracy, which significantly improves the 3D object detection performance. Experimental results on the widely used KITTI dataset and the newly released large-scale Waymo Open Dataset demonstrate that our method achieves better performance than the anchor-based 3D object detectors, and each individual component in our method is effective with significant performance gains.

\begin{acks}
This work was supported by the National Key Research and Development Program of China under Grant 2018YFE0183900.
\end{acks}


\begin{thebibliography}{68}


\ifx \showCODEN    \undefined \def \showCODEN     #1{\unskip}     \fi
\ifx \showDOI      \undefined \def \showDOI       #1{#1}\fi
\ifx \showISBNx    \undefined \def \showISBNx     #1{\unskip}     \fi
\ifx \showISBNxiii \undefined \def \showISBNxiii  #1{\unskip}     \fi
\ifx \showISSN     \undefined \def \showISSN      #1{\unskip}     \fi
\ifx \showLCCN     \undefined \def \showLCCN      #1{\unskip}     \fi
\ifx \shownote     \undefined \def \shownote      #1{#1}          \fi
\ifx \showarticletitle \undefined \def \showarticletitle #1{#1}   \fi
\ifx \showURL      \undefined \def \showURL       {\relax}        \fi
\providecommand\bibfield[2]{#2}
\providecommand\bibinfo[2]{#2}
\providecommand\natexlab[1]{#1}
\providecommand\showeprint[2][]{arXiv:#2}

\bibitem[\protect\citeauthoryear{Cai and Vasconcelos}{Cai and
  Vasconcelos}{2018}]%
        {Cascadercnn}
\bibfield{author}{\bibinfo{person}{Zhaowei Cai} {and} \bibinfo{person}{Nuno
  Vasconcelos}.} \bibinfo{year}{2018}\natexlab{}.
\newblock \showarticletitle{Cascade {R-CNN}: Delving Into High Quality Object
  Detection}. In \bibinfo{booktitle}{\emph{2018 IEEE/CVF Conference on Computer
  Vision and Pattern Recognition (CVPR)}}. \bibinfo{pages}{6154--6162}.
\newblock


\bibitem[\protect\citeauthoryear{Chen, Sun, Wang, Jia, and Yuille}{Chen
  et~al\mbox{.}}{2020}]%
        {Hotspots(ECCV2020)}
\bibfield{author}{\bibinfo{person}{Qi Chen}, \bibinfo{person}{Lin Sun},
  \bibinfo{person}{Zhixin Wang}, \bibinfo{person}{Kui Jia}, {and}
  \bibinfo{person}{Alan Yuille}.} \bibinfo{year}{2020}\natexlab{}.
\newblock \showarticletitle{Object as Hotspots: An Anchor-Free {3D} Object
  Detection Approach via Firing of Hotspots}. In
  \bibinfo{booktitle}{\emph{European Conference on Computer Vision (ECCV)}}.
  \bibinfo{pages}{68--84}.
\newblock
\showISBNx{978-3-030-58589-1}


\bibitem[\protect\citeauthoryear{Chen, Ma, Wan, Li, and Xia}{Chen
  et~al\mbox{.}}{2017}]%
        {MV3D}
\bibfield{author}{\bibinfo{person}{Xiaozhi Chen}, \bibinfo{person}{Huimin Ma},
  \bibinfo{person}{Ji Wan}, \bibinfo{person}{Bo Li}, {and}
  \bibinfo{person}{Tian Xia}.} \bibinfo{year}{2017}\natexlab{}.
\newblock \showarticletitle{Multi-view {3D} Object Detection Network for
  Autonomous Driving}. In \bibinfo{booktitle}{\emph{2017 IEEE Conference on
  Computer Vision and Pattern Recognition (CVPR)}}.
  \bibinfo{pages}{6526--6534}.
\newblock
\showISSN{1063-6919}


\bibitem[\protect\citeauthoryear{Chen, Liu, Shen, and Jia}{Chen
  et~al\mbox{.}}{2019}]%
        {Fast-PointRCNN}
\bibfield{author}{\bibinfo{person}{Yilun Chen}, \bibinfo{person}{Shu Liu},
  \bibinfo{person}{Xiaoyong Shen}, {and} \bibinfo{person}{Jiaya Jia}.}
  \bibinfo{year}{2019}\natexlab{}.
\newblock \showarticletitle{Fast Point {R-CNN}}. In
  \bibinfo{booktitle}{\emph{2019 {IEEE/CVF} International Conference on
  Computer Vision (ICCV)}}. \bibinfo{pages}{9775--9784}.
\newblock


\bibitem[\protect\citeauthoryear{Dai, Luo, Ding, and Shao}{Dai
  et~al\mbox{.}}{2020}]%
        {dai2020commands}
\bibfield{author}{\bibinfo{person}{Hang Dai}, \bibinfo{person}{Shujie Luo},
  \bibinfo{person}{Yong Ding}, {and} \bibinfo{person}{Ling Shao}.}
  \bibinfo{year}{2020}\natexlab{}.
\newblock \showarticletitle{Commands for autonomous vehicles by progressively
  stacking visual-linguistic representations}. In
  \bibinfo{booktitle}{\emph{European Conference on Computer Vision (ECCV)}}.
  Springer, \bibinfo{pages}{27--32}.
\newblock


\bibitem[\protect\citeauthoryear{Dai, Qi, Xiong, Li, Zhang, Hu, and Wei}{Dai
  et~al\mbox{.}}{2017}]%
        {DCN}
\bibfield{author}{\bibinfo{person}{Jifeng Dai}, \bibinfo{person}{Haozhi Qi},
  \bibinfo{person}{Yuwen Xiong}, \bibinfo{person}{Yi Li},
  \bibinfo{person}{Guodong Zhang}, \bibinfo{person}{Han Hu}, {and}
  \bibinfo{person}{Yichen Wei}.} \bibinfo{year}{2017}\natexlab{}.
\newblock \showarticletitle{Deformable Convolutional Networks}. In
  \bibinfo{booktitle}{\emph{2017 IEEE International Conference on Computer
  Vision (ICCV)}}. \bibinfo{pages}{764--773}.
\newblock


\bibitem[\protect\citeauthoryear{Du, Ye, Tan, Feng, Xu, Ding, and Wen}{Du
  et~al\mbox{.}}{2020}]%
        {Associate-3Ddet}
\bibfield{author}{\bibinfo{person}{Liang Du}, \bibinfo{person}{Xiaoqing Ye},
  \bibinfo{person}{Xiao Tan}, \bibinfo{person}{Jianfeng Feng},
  \bibinfo{person}{Zhenbo Xu}, \bibinfo{person}{Errui Ding}, {and}
  \bibinfo{person}{Shilei Wen}.} \bibinfo{year}{2020}\natexlab{}.
\newblock \showarticletitle{{Associate-3Ddet}: Perceptual-to-Conceptual
  Association for {3D} Point Cloud Object Detection}. In
  \bibinfo{booktitle}{\emph{2020 {IEEE/CVF} Conference on Computer Vision and
  Pattern Recognition (CVPR)}}. \bibinfo{pages}{13326--13335}.
\newblock


\bibitem[\protect\citeauthoryear{Duan, Bai, Xie, Qi, Huang, and Tian}{Duan
  et~al\mbox{.}}{2019}]%
        {CornerNet_triplets}
\bibfield{author}{\bibinfo{person}{Kaiwen Duan}, \bibinfo{person}{Song Bai},
  \bibinfo{person}{Lingxi Xie}, \bibinfo{person}{Honggang Qi},
  \bibinfo{person}{Qingming Huang}, {and} \bibinfo{person}{Qi Tian}.}
  \bibinfo{year}{2019}\natexlab{}.
\newblock \showarticletitle{{CenterNet}: Keypoint Triplets for Object
  Detection}. In \bibinfo{booktitle}{\emph{2019 IEEE/CVF International
  Conference on Computer Vision (ICCV)}}. \bibinfo{pages}{6568--6577}.
\newblock


\bibitem[\protect\citeauthoryear{Fei, Chen, Heidenreich, Wirges, and
  Stiller}{Fei et~al\mbox{.}}{2020}]%
        {semanticvoxels}
\bibfield{author}{\bibinfo{person}{Juncong Fei}, \bibinfo{person}{Wenbo Chen},
  \bibinfo{person}{Philipp Heidenreich}, \bibinfo{person}{Sascha Wirges}, {and}
  \bibinfo{person}{Christoph Stiller}.} \bibinfo{year}{2020}\natexlab{}.
\newblock \showarticletitle{SemanticVoxels: Sequential Fusion for 3D Pedestrian
  Detection using LiDAR Point Cloud and Semantic Segmentation}. In
  \bibinfo{booktitle}{\emph{{IEEE} International Conference on Multisensor
  Fusion and Integration for Intelligent Systems}}.
  \bibinfo{publisher}{{IEEE}}, \bibinfo{pages}{185--190}.
\newblock


\bibitem[\protect\citeauthoryear{Feng, Jiao, Zhu, and Sun}{Feng
  et~al\mbox{.}}{2020}]%
        {otracking_acmmm20}
\bibfield{author}{\bibinfo{person}{Tuo Feng}, \bibinfo{person}{Licheng Jiao},
  \bibinfo{person}{Hao Zhu}, {and} \bibinfo{person}{Long Sun}.}
  \bibinfo{year}{2020}\natexlab{}.
\newblock \showarticletitle{A Novel Object Re-Track Framework for 3D Point
  Clouds}. In \bibinfo{booktitle}{\emph{{MM} '20: The 28th {ACM} International
  Conference on Multimedia}}. \bibinfo{publisher}{{ACM}},
  \bibinfo{pages}{3118--3126}.
\newblock


\bibitem[\protect\citeauthoryear{Geiger, Lenz, and Urtasun}{Geiger
  et~al\mbox{.}}{2012}]%
        {KITTIDataset}
\bibfield{author}{\bibinfo{person}{Andreas Geiger}, \bibinfo{person}{Philip
  Lenz}, {and} \bibinfo{person}{Raquel Urtasun}.}
  \bibinfo{year}{2012}\natexlab{}.
\newblock \showarticletitle{Are We Ready for Autonomous Driving? The {KITTI}
  Vision Benchmark Suite}. In \bibinfo{booktitle}{\emph{2012 {IEEE} Conference
  on Computer Vision and Pattern Recognition (CVPR)}}.
  \bibinfo{pages}{3354--3361}.
\newblock
\showISSN{1063-6919}


\bibitem[\protect\citeauthoryear{Graham}{Graham}{2015}]%
        {sparseconv}
\bibfield{author}{\bibinfo{person}{Ben Graham}.}
  \bibinfo{year}{2015}\natexlab{}.
\newblock \showarticletitle{Sparse {3D} Convolutional Neural Networks}. In
  \bibinfo{booktitle}{\emph{2015 British Machine Vision Conference (BMVC)}}.
  \bibinfo{pages}{150.1--150.9}.
\newblock


\bibitem[\protect\citeauthoryear{Graham, Engelcke, and van~der Maaten}{Graham
  et~al\mbox{.}}{2018}]%
        {submanifold}
\bibfield{author}{\bibinfo{person}{Benjamin Graham}, \bibinfo{person}{Martin
  Engelcke}, {and} \bibinfo{person}{Laurens van~der Maaten}.}
  \bibinfo{year}{2018}\natexlab{}.
\newblock \showarticletitle{{3D} Semantic Segmentation With Submanifold Sparse
  Convolutional Networks}. In \bibinfo{booktitle}{\emph{2018 IEEE/CVF
  Conference on Computer Vision and Pattern Recognition (CVPR)}}.
  \bibinfo{pages}{9224--9232}.
\newblock
\showISSN{1063-6919}


\bibitem[\protect\citeauthoryear{He, Zeng, Huang, Hua, and Zhang}{He
  et~al\mbox{.}}{2020}]%
        {SASSD}
\bibfield{author}{\bibinfo{person}{Chenhang He}, \bibinfo{person}{Hui Zeng},
  \bibinfo{person}{Jianqiang Huang}, \bibinfo{person}{Xian{-}Sheng Hua}, {and}
  \bibinfo{person}{Lei Zhang}.} \bibinfo{year}{2020}\natexlab{}.
\newblock \showarticletitle{Structure Aware Single-Stage {3D} Object Detection
  From Point Cloud}. In \bibinfo{booktitle}{\emph{2020 IEEE/CVF Conference on
  Computer Vision and Pattern Recognition (CVPR)}}.
  \bibinfo{pages}{11870--11879}.
\newblock


\bibitem[\protect\citeauthoryear{He, Gkioxari, Doll{\'{a}}r, and Girshick}{He
  et~al\mbox{.}}{2017}]%
        {Maskrcnn}
\bibfield{author}{\bibinfo{person}{Kaiming He}, \bibinfo{person}{Georgia
  Gkioxari}, \bibinfo{person}{Piotr Doll{\'{a}}r}, {and}
  \bibinfo{person}{Ross~B. Girshick}.} \bibinfo{year}{2017}\natexlab{}.
\newblock \showarticletitle{Mask {R-CNN}}. In \bibinfo{booktitle}{\emph{2017
  IEEE International Conference on Computer Vision (ICCV)}}.
  \bibinfo{pages}{2980--2988}.
\newblock


\bibitem[\protect\citeauthoryear{He, Zhang, Ren, and Sun}{He
  et~al\mbox{.}}{2016}]%
        {ResNet}
\bibfield{author}{\bibinfo{person}{Kaiming He}, \bibinfo{person}{Xiangyu
  Zhang}, \bibinfo{person}{Shaoqing Ren}, {and} \bibinfo{person}{Jian Sun}.}
  \bibinfo{year}{2016}\natexlab{}.
\newblock \showarticletitle{Deep Residual Learning for Image Recognition}. In
  \bibinfo{booktitle}{\emph{2016 IEEE Conference on Computer Vision and Pattern
  Recognition (CVPR)}}. \bibinfo{pages}{770--778}.
\newblock


\bibitem[\protect\citeauthoryear{Hsu}{Hsu}{2019}]%
        {learningfrom3d_acmmm19}
\bibfield{author}{\bibinfo{person}{Winston~H. Hsu}.}
  \bibinfo{year}{2019}\natexlab{}.
\newblock \showarticletitle{Learning from 3D (Point Cloud) Data}. In
  \bibinfo{booktitle}{\emph{{MM} '19: The 27th {ACM} International Conference
  on Multimedia}}. \bibinfo{publisher}{{ACM}}, \bibinfo{pages}{2697--2698}.
\newblock


\bibitem[\protect\citeauthoryear{Jiang, Luo, Mao, Xiao, and Jiang}{Jiang
  et~al\mbox{.}}{2018}]%
        {IoUNet}
\bibfield{author}{\bibinfo{person}{Borui Jiang}, \bibinfo{person}{Ruixuan Luo},
  \bibinfo{person}{Jiayuan Mao}, \bibinfo{person}{Tete Xiao}, {and}
  \bibinfo{person}{Yuning Jiang}.} \bibinfo{year}{2018}\natexlab{}.
\newblock \showarticletitle{Acquisition of Localization Confidence for Accurate
  Object Detection}. In \bibinfo{booktitle}{\emph{European Conference on
  Computer Vision (ECCV)}}. \bibinfo{pages}{784--799}.
\newblock


\bibitem[\protect\citeauthoryear{KITTI}{KITTI}{2021}]%
        {KITTIleaderboard}
\bibfield{author}{\bibinfo{person}{KITTI}.} \bibinfo{year}{2021}\natexlab{}.
\newblock \bibinfo{title}{{KITTI} Leaderboard of {3D} Object Detection
  Benchmark}.
\newblock
\newblock
\urldef\tempurl%
\url{http://www.cvlibs.net/datasets/kitti/eval_object.php?obj_benchmark=3d}
\showURL{%
\tempurl}


\bibitem[\protect\citeauthoryear{Ku, Mozifian, Lee, Harakeh, and Waslander}{Ku
  et~al\mbox{.}}{2018}]%
        {AVOD}
\bibfield{author}{\bibinfo{person}{Jason Ku}, \bibinfo{person}{Melissa
  Mozifian}, \bibinfo{person}{Jungwook Lee}, \bibinfo{person}{Ali Harakeh},
  {and} \bibinfo{person}{Steven~L. Waslander}.}
  \bibinfo{year}{2018}\natexlab{}.
\newblock \showarticletitle{Joint {3D} Proposal Generation and Object Detection
  from View Aggregation}. In \bibinfo{booktitle}{\emph{2018 IEEE/RSJ
  International Conference on Intelligent Robots and Systems (IROS)}}.
  \bibinfo{pages}{1--8}.
\newblock
\showISSN{2153-0858}


\bibitem[\protect\citeauthoryear{Lang, Vora, Caesar, Zhou, Yang, and
  Beijbom}{Lang et~al\mbox{.}}{2019}]%
        {PointPillars}
\bibfield{author}{\bibinfo{person}{Alex~H. Lang}, \bibinfo{person}{Sourabh
  Vora}, \bibinfo{person}{Holger Caesar}, \bibinfo{person}{Lubing Zhou},
  \bibinfo{person}{Jiong Yang}, {and} \bibinfo{person}{Oscar Beijbom}.}
  \bibinfo{year}{2019}\natexlab{}.
\newblock \showarticletitle{{PointPillars}: Fast Encoders for Object Detection
  From Point Clouds}. In \bibinfo{booktitle}{\emph{2019 IEEE/CVF Conference on
  Computer Vision and Pattern Recognition (CVPR)}}.
  \bibinfo{pages}{12689--12697}.
\newblock
\showISSN{1063-6919}


\bibitem[\protect\citeauthoryear{Law and Deng}{Law and Deng}{2018}]%
        {Cornernet}
\bibfield{author}{\bibinfo{person}{Hei Law} {and} \bibinfo{person}{Jia Deng}.}
  \bibinfo{year}{2018}\natexlab{}.
\newblock \showarticletitle{CornerNet: Detecting Objects as Paired Keypoints}.
  In \bibinfo{booktitle}{\emph{European Conference on Computer Vision (ECCV)}},
  Vol.~\bibinfo{volume}{11218}. \bibinfo{pages}{765--781}.
\newblock


\bibitem[\protect\citeauthoryear{Li, Luo, Zhu, Dai, Krylov, Ding, and Shao}{Li
  et~al\mbox{.}}{2020a}]%
        {li20203d}
\bibfield{author}{\bibinfo{person}{Jiale Li}, \bibinfo{person}{Shujie Luo},
  \bibinfo{person}{Ziqi Zhu}, \bibinfo{person}{Hang Dai},
  \bibinfo{person}{Andrey~S Krylov}, \bibinfo{person}{Yong Ding}, {and}
  \bibinfo{person}{Ling Shao}.} \bibinfo{year}{2020}\natexlab{a}.
\newblock \showarticletitle{3D IoU-Net: IoU guided {3D} object detector for
  point clouds}.
\newblock \bibinfo{journal}{\emph{arXiv preprint arXiv:2004.04962}}
  (\bibinfo{year}{2020}).
\newblock


\bibitem[\protect\citeauthoryear{Li, Sun, Luo, Zhu, Dai, Krylov, Ding, and
  Shao}{Li et~al\mbox{.}}{2021}]%
        {li2021p2v}
\bibfield{author}{\bibinfo{person}{Jiale Li}, \bibinfo{person}{Yu Sun},
  \bibinfo{person}{Shujie Luo}, \bibinfo{person}{Ziqi Zhu},
  \bibinfo{person}{Hang Dai}, \bibinfo{person}{Andrey~S Krylov},
  \bibinfo{person}{Yong Ding}, {and} \bibinfo{person}{Ling Shao}.}
  \bibinfo{year}{2021}\natexlab{}.
\newblock \showarticletitle{P2V-RCNN: Point to Voxel Feature Learning for 3D
  Object Detection from Point Clouds}.
\newblock \bibinfo{journal}{\emph{IEEE Access}} (\bibinfo{year}{2021}).
\newblock


\bibitem[\protect\citeauthoryear{Li, Wang, Wu, Chen, Hu, Li, Tang, and Yang}{Li
  et~al\mbox{.}}{2020b}]%
        {GFocal}
\bibfield{author}{\bibinfo{person}{Xiang Li}, \bibinfo{person}{Wenhai Wang},
  \bibinfo{person}{Lijun Wu}, \bibinfo{person}{Shuo Chen},
  \bibinfo{person}{Xiaolin Hu}, \bibinfo{person}{Jun Li},
  \bibinfo{person}{Jinhui Tang}, {and} \bibinfo{person}{Jian Yang}.}
  \bibinfo{year}{2020}\natexlab{b}.
\newblock \showarticletitle{Generalized Focal Loss: Learning Qualified and
  Distributed Bounding Boxes for Dense Object Detection}. In
  \bibinfo{booktitle}{\emph{Conference on Neural Information Processing Systems
  (NeurIPS)}}.
\newblock


\bibitem[\protect\citeauthoryear{Liang, Yang, Chen, Hu, and Urtasun}{Liang
  et~al\mbox{.}}{2019}]%
        {MMF}
\bibfield{author}{\bibinfo{person}{Ming Liang}, \bibinfo{person}{Bin Yang},
  \bibinfo{person}{Yun Chen}, \bibinfo{person}{Rui Hu}, {and}
  \bibinfo{person}{Raquel Urtasun}.} \bibinfo{year}{2019}\natexlab{}.
\newblock \showarticletitle{Multi-Task Multi-Sensor Fusion for {3D} Object
  Detection}. In \bibinfo{booktitle}{\emph{2019 IEEE/CVF Conference on Computer
  Vision and Pattern Recognition (CVPR)}}. \bibinfo{pages}{7337--7345}.
\newblock
\showISSN{1063-6919}


\bibitem[\protect\citeauthoryear{Lin, Doll{\'{a}}r, Girshick, He, Hariharan,
  and Belongie}{Lin et~al\mbox{.}}{2017}]%
        {FPN}
\bibfield{author}{\bibinfo{person}{Tsung{-}Yi Lin}, \bibinfo{person}{Piotr
  Doll{\'{a}}r}, \bibinfo{person}{Ross~B. Girshick}, \bibinfo{person}{Kaiming
  He}, \bibinfo{person}{Bharath Hariharan}, {and} \bibinfo{person}{Serge~J.
  Belongie}.} \bibinfo{year}{2017}\natexlab{}.
\newblock \showarticletitle{Feature Pyramid Networks for Object Detection}. In
  \bibinfo{booktitle}{\emph{2017 IEEE Conference on Computer Vision and Pattern
  Recognition (CVPR)}}. \bibinfo{pages}{936--944}.
\newblock


\bibitem[\protect\citeauthoryear{Lin, Goyal, Girshick, He, and
  Doll{\'{a}}r}{Lin et~al\mbox{.}}{2020}]%
        {RetinaNet}
\bibfield{author}{\bibinfo{person}{Tsung{-}Yi Lin}, \bibinfo{person}{Priya
  Goyal}, \bibinfo{person}{Ross~B. Girshick}, \bibinfo{person}{Kaiming He},
  {and} \bibinfo{person}{Piotr Doll{\'{a}}r}.} \bibinfo{year}{2020}\natexlab{}.
\newblock \showarticletitle{Focal Loss for Dense Object Detection}.
\newblock \bibinfo{journal}{\emph{IEEE Transactions on Pattern Analysis and
  Machine Intelligence}} \bibinfo{volume}{42}, \bibinfo{number}{2}
  (\bibinfo{year}{2020}), \bibinfo{pages}{318--327}.
\newblock


\bibitem[\protect\citeauthoryear{Liu, Anguelov, Erhan, Szegedy, Reed, Fu, and
  Berg}{Liu et~al\mbox{.}}{2016}]%
        {ssd}
\bibfield{author}{\bibinfo{person}{Wei Liu}, \bibinfo{person}{Dragomir
  Anguelov}, \bibinfo{person}{Dumitru Erhan}, \bibinfo{person}{Christian
  Szegedy}, \bibinfo{person}{Scott~E. Reed}, \bibinfo{person}{Cheng{-}Yang Fu},
  {and} \bibinfo{person}{Alexander~C. Berg}.} \bibinfo{year}{2016}\natexlab{}.
\newblock \showarticletitle{{SSD}: Single Shot Multibox Detector}. In
  \bibinfo{booktitle}{\emph{European Conference on Computer Vision (ECCV)}}.
  Springer, \bibinfo{pages}{21--37}.
\newblock


\bibitem[\protect\citeauthoryear{Liu, Zhao, Huang, Hu, Zhou, and Bai}{Liu
  et~al\mbox{.}}{2020}]%
        {tanet_AAAI}
\bibfield{author}{\bibinfo{person}{Zhe Liu}, \bibinfo{person}{Xin Zhao},
  \bibinfo{person}{Tengteng Huang}, \bibinfo{person}{Ruolan Hu},
  \bibinfo{person}{Yu Zhou}, {and} \bibinfo{person}{Xiang Bai}.}
  \bibinfo{year}{2020}\natexlab{}.
\newblock \showarticletitle{{TANet}: Robust {3D} Object Detection from Point
  Clouds with Triple Attention}. In \bibinfo{booktitle}{\emph{2020 {AAAI}
  Conference on Artificial Intelligence (AAAI)}}.
  \bibinfo{pages}{11677--11684}.
\newblock


\bibitem[\protect\citeauthoryear{Loshchilov and Hutter}{Loshchilov and
  Hutter}{2017}]%
        {AdamW}
\bibfield{author}{\bibinfo{person}{Ilya Loshchilov} {and}
  \bibinfo{person}{Frank Hutter}.} \bibinfo{year}{2017}\natexlab{}.
\newblock \showarticletitle{Fixing Weight Decay Regularization in Adam}.
\newblock \bibinfo{journal}{\emph{CoRR}}  \bibinfo{volume}{abs/1711.05101}
  (\bibinfo{year}{2017}).
\newblock
\showeprint[arxiv]{1711.05101}


\bibitem[\protect\citeauthoryear{Luo, Dai, Shao, and Ding}{Luo
  et~al\mbox{.}}{2020}]%
        {luo2020c4av}
\bibfield{author}{\bibinfo{person}{Shujie Luo}, \bibinfo{person}{Hang Dai},
  \bibinfo{person}{Ling Shao}, {and} \bibinfo{person}{Yong Ding}.}
  \bibinfo{year}{2020}\natexlab{}.
\newblock \showarticletitle{C4AV: Learning Cross-Modal Representations from
  Transformers}. In \bibinfo{booktitle}{\emph{European Conference on Computer
  Vision (ECCV)}}. Springer, \bibinfo{pages}{33--38}.
\newblock


\bibitem[\protect\citeauthoryear{Luo, Dai, Shao, and Ding}{Luo
  et~al\mbox{.}}{2021}]%
        {luo2021m3dssd}
\bibfield{author}{\bibinfo{person}{Shujie Luo}, \bibinfo{person}{Hang Dai},
  \bibinfo{person}{Ling Shao}, {and} \bibinfo{person}{Yong Ding}.}
  \bibinfo{year}{2021}\natexlab{}.
\newblock \showarticletitle{M3DSSD: Monocular 3D single stage object detector}.
  In \bibinfo{booktitle}{\emph{IEEE/CVF Conference on Computer Vision and
  Pattern Recognition (CVPR)}}. \bibinfo{pages}{6145--6154}.
\newblock


\bibitem[\protect\citeauthoryear{Pang, Morris, and Radha}{Pang
  et~al\mbox{.}}{2020}]%
        {CLOCs}
\bibfield{author}{\bibinfo{person}{Su Pang}, \bibinfo{person}{Daniel~D.
  Morris}, {and} \bibinfo{person}{Hayder Radha}.}
  \bibinfo{year}{2020}\natexlab{}.
\newblock \showarticletitle{CLOCs: Camera-LiDAR Object Candidates Fusion for 3D
  Object Detection}. In \bibinfo{booktitle}{\emph{{IEEE/RSJ} International
  Conference on Intelligent Robots and Systems (IROS)}}.
  \bibinfo{pages}{10386--10393}.
\newblock


\bibitem[\protect\citeauthoryear{Qi, Liu, Wu, Su, and Guibas}{Qi
  et~al\mbox{.}}{2018}]%
        {F-PointNets}
\bibfield{author}{\bibinfo{person}{Charles~R. Qi}, \bibinfo{person}{Wei Liu},
  \bibinfo{person}{Chenxia Wu}, \bibinfo{person}{Hao Su}, {and}
  \bibinfo{person}{Leonidas~J. Guibas}.} \bibinfo{year}{2018}\natexlab{}.
\newblock \showarticletitle{Frustum {PointNets} for {3D} Object Detection from
  {RGB-D} Data}. In \bibinfo{booktitle}{\emph{2018 IEEE/CVF Conference on
  Computer Vision and Pattern Recognition (CVPR)}}. \bibinfo{pages}{918--927}.
\newblock
\showISSN{1063-6919}


\bibitem[\protect\citeauthoryear{Qi, Su, Mo, and Guibas}{Qi
  et~al\mbox{.}}{2017a}]%
        {PointNet}
\bibfield{author}{\bibinfo{person}{Charles~Ruizhongtai Qi},
  \bibinfo{person}{Hao Su}, \bibinfo{person}{Kaichun Mo}, {and}
  \bibinfo{person}{Leonidas~J. Guibas}.} \bibinfo{year}{2017}\natexlab{a}.
\newblock \showarticletitle{{PointNet}: Deep Learning on Point Sets for {3D}
  Classification and Segmentation}. In \bibinfo{booktitle}{\emph{2017 IEEE
  Conference on Computer Vision and Pattern Recognition (CVPR)}}.
  \bibinfo{pages}{77--85}.
\newblock
\showISSN{1063-6919}


\bibitem[\protect\citeauthoryear{Qi, Yi, Su, and Guibas}{Qi
  et~al\mbox{.}}{2017b}]%
        {PointNet++}
\bibfield{author}{\bibinfo{person}{Charles~Ruizhongtai Qi}, \bibinfo{person}{Li
  Yi}, \bibinfo{person}{Hao Su}, {and} \bibinfo{person}{Leonidas~J. Guibas}.}
  \bibinfo{year}{2017}\natexlab{b}.
\newblock \showarticletitle{{PointNet}++: Deep Hierarchical Feature Learning on
  Point Sets in a Metric Space}. In \bibinfo{booktitle}{\emph{Conference on
  Neural Information Processing Systems (NeurIPS)}}.
  \bibinfo{pages}{5099--5108}.
\newblock


\bibitem[\protect\citeauthoryear{Qin, Wang, and Lu}{Qin et~al\mbox{.}}{2020}]%
        {WeaklySup_PC_3DOD_acmmm20}
\bibfield{author}{\bibinfo{person}{Zengyi Qin}, \bibinfo{person}{Jinglu Wang},
  {and} \bibinfo{person}{Yan Lu}.} \bibinfo{year}{2020}\natexlab{}.
\newblock \showarticletitle{Weakly Supervised {3D} Object Detection from Point
  Clouds}. In \bibinfo{booktitle}{\emph{{MM} '20: The 28th {ACM} International
  Conference on Multimedia}}. \bibinfo{publisher}{{ACM}},
  \bibinfo{pages}{4144--4152}.
\newblock


\bibitem[\protect\citeauthoryear{Redmon, Divvala, Girshick, and Farhadi}{Redmon
  et~al\mbox{.}}{2016}]%
        {YOLO}
\bibfield{author}{\bibinfo{person}{Joseph Redmon},
  \bibinfo{person}{Santosh~Kumar Divvala}, \bibinfo{person}{Ross~B. Girshick},
  {and} \bibinfo{person}{Ali Farhadi}.} \bibinfo{year}{2016}\natexlab{}.
\newblock \showarticletitle{You Only Look Once: Unified, Real-Time Object
  Detection}. In \bibinfo{booktitle}{\emph{2016 IEEE Conference on Computer
  Vision and Pattern Recognition (CVPR)}}. \bibinfo{pages}{779--788}.
\newblock


\bibitem[\protect\citeauthoryear{Ren, He, Girshick, and Sun}{Ren
  et~al\mbox{.}}{2017}]%
        {fasterrcnn}
\bibfield{author}{\bibinfo{person}{Shaoqing Ren}, \bibinfo{person}{Kaiming He},
  \bibinfo{person}{Ross~B. Girshick}, {and} \bibinfo{person}{Jian Sun}.}
  \bibinfo{year}{2017}\natexlab{}.
\newblock \showarticletitle{Faster {R-CNN}: Towards Real-Time Object Detection
  with Region Proposal Networks}.
\newblock \bibinfo{journal}{\emph{IEEE Transactions on Pattern Analysis and
  Machine Intelligence}} \bibinfo{volume}{39}, \bibinfo{number}{6}
  (\bibinfo{year}{2017}), \bibinfo{pages}{1137--1149}.
\newblock


\bibitem[\protect\citeauthoryear{Saputra, Rakicevic, and Kormushev}{Saputra
  et~al\mbox{.}}{2019}]%
        {PC_ROBOTS}
\bibfield{author}{\bibinfo{person}{Roni~Permana Saputra},
  \bibinfo{person}{Nemanja Rakicevic}, {and} \bibinfo{person}{Petar
  Kormushev}.} \bibinfo{year}{2019}\natexlab{}.
\newblock \showarticletitle{Sim-to-Real Learning for Casualty Detection from
  Ground Projected Point Cloud Data}. In \bibinfo{booktitle}{\emph{2019
  IEEE/RSJ International Conference on Intelligent Robots and Systems (IROS)}}.
  \bibinfo{pages}{3918--3925}.
\newblock


\bibitem[\protect\citeauthoryear{Shi, Guo, Jiang, Wang, Shi, Wang, and Li}{Shi
  et~al\mbox{.}}{2020a}]%
        {PVRCNN}
\bibfield{author}{\bibinfo{person}{Shaoshuai Shi}, \bibinfo{person}{Chaoxu
  Guo}, \bibinfo{person}{Li Jiang}, \bibinfo{person}{Zhe Wang},
  \bibinfo{person}{Jianping Shi}, \bibinfo{person}{Xiaogang Wang}, {and}
  \bibinfo{person}{Hongsheng Li}.} \bibinfo{year}{2020}\natexlab{a}.
\newblock \showarticletitle{{PV-RCNN}: Point-Voxel Feature Set Abstraction for
  {3D} Object Detection}. In \bibinfo{booktitle}{\emph{2020 IEEE/CVF Conference
  on Computer Vision and Pattern Recognition (CVPR)}}.
  \bibinfo{pages}{10526--10535}.
\newblock


\bibitem[\protect\citeauthoryear{Shi, Wang, and Li}{Shi et~al\mbox{.}}{2019}]%
        {PointRCNN}
\bibfield{author}{\bibinfo{person}{Shaoshuai Shi}, \bibinfo{person}{Xiaogang
  Wang}, {and} \bibinfo{person}{Hongsheng Li}.}
  \bibinfo{year}{2019}\natexlab{}.
\newblock \showarticletitle{{PointRCNN}: {3D} Object Proposal Generation and
  Detection From Point Cloud}. In \bibinfo{booktitle}{\emph{2019 IEEE/CVF
  Conference on Computer Vision and Pattern Recognition (CVPR)}}.
  \bibinfo{pages}{770--779}.
\newblock
\showISSN{1063-6919}


\bibitem[\protect\citeauthoryear{Shi, Wang, Shi, Wang, and Li}{Shi
  et~al\mbox{.}}{2020b}]%
        {PartA2_TPAMI}
\bibfield{author}{\bibinfo{person}{Shaoshuai Shi}, \bibinfo{person}{Zhe Wang},
  \bibinfo{person}{Jianping Shi}, \bibinfo{person}{Xiaogang Wang}, {and}
  \bibinfo{person}{Hongsheng Li}.} \bibinfo{year}{2020}\natexlab{b}.
\newblock \showarticletitle{From Points to Parts: {3D} Object Detection from
  Point Cloud with Part-aware and Part-aggregation Network}.
\newblock \bibinfo{journal}{\emph{IEEE Transactions on Pattern Analysis and
  Machine Intelligence}} (\bibinfo{year}{2020}), \bibinfo{pages}{1--1}.
\newblock


\bibitem[\protect\citeauthoryear{Shi and Rajkumar}{Shi and Rajkumar}{2020}]%
        {PointGNN}
\bibfield{author}{\bibinfo{person}{Weijing Shi} {and} \bibinfo{person}{Raj
  Rajkumar}.} \bibinfo{year}{2020}\natexlab{}.
\newblock \showarticletitle{{Point-GNN}: Graph Neural Network for {3D} Object
  Detection in a Point Cloud}. In \bibinfo{booktitle}{\emph{2020 IEEE/CVF
  Conference on Computer Vision and Pattern Recognition (CVPR)}}.
  \bibinfo{pages}{1708--1716}.
\newblock


\bibitem[\protect\citeauthoryear{Smith and Topin}{Smith and Topin}{2019}]%
        {one_cycle_lr}
\bibfield{author}{\bibinfo{person}{Leslie~N. Smith} {and}
  \bibinfo{person}{Nicholay Topin}.} \bibinfo{year}{2019}\natexlab{}.
\newblock \showarticletitle{Super-convergence: Very Fast Training of Neural
  Networks Using Large Learning Rates}. In \bibinfo{booktitle}{\emph{Artificial
  Intelligence and Machine Learning for Multi-Domain Operations Applications}},
  Vol.~\bibinfo{volume}{11006}. \bibinfo{pages}{1100612}.
\newblock


\bibitem[\protect\citeauthoryear{Sun, Kretzschmar, Dotiwalla, Chouard, Patnaik,
  Tsui, Guo, Zhou, Chai, Caine, Vasudevan, Han, Ngiam, Zhao, Timofeev,
  Ettinger, Krivokon, Gao, Joshi, Zhang, Shlens, Chen, and Anguelov}{Sun
  et~al\mbox{.}}{2020a}]%
        {waymo_open_dataset}
\bibfield{author}{\bibinfo{person}{Pei Sun}, \bibinfo{person}{Henrik
  Kretzschmar}, \bibinfo{person}{Xerxes Dotiwalla}, \bibinfo{person}{Aurelien
  Chouard}, \bibinfo{person}{Vijaysai Patnaik}, \bibinfo{person}{Paul Tsui},
  \bibinfo{person}{James Guo}, \bibinfo{person}{Yin Zhou},
  \bibinfo{person}{Yuning Chai}, \bibinfo{person}{Benjamin Caine},
  \bibinfo{person}{Vijay Vasudevan}, \bibinfo{person}{Wei Han},
  \bibinfo{person}{Jiquan Ngiam}, \bibinfo{person}{Hang Zhao},
  \bibinfo{person}{Aleksei Timofeev}, \bibinfo{person}{Scott Ettinger},
  \bibinfo{person}{Maxim Krivokon}, \bibinfo{person}{Amy Gao},
  \bibinfo{person}{Aditya Joshi}, \bibinfo{person}{Yu Zhang},
  \bibinfo{person}{Jonathon Shlens}, \bibinfo{person}{Zhifeng Chen}, {and}
  \bibinfo{person}{Dragomir Anguelov}.} \bibinfo{year}{2020}\natexlab{a}.
\newblock \showarticletitle{Scalability in Perception for Autonomous Driving:
  Waymo Open Dataset}. In \bibinfo{booktitle}{\emph{2020 {IEEE/CVF} Conference
  on Computer Vision and Pattern Recognition (CVPR)}}.
  \bibinfo{pages}{2443--2451}.
\newblock


\bibitem[\protect\citeauthoryear{Sun, Wang, Wang, Cheng, and Liu}{Sun
  et~al\mbox{.}}{2020b}]%
        {rang_img_pointcloud}
\bibfield{author}{\bibinfo{person}{Xuebin Sun}, \bibinfo{person}{Sukai Wang},
  \bibinfo{person}{Miaohui Wang}, \bibinfo{person}{Shing~Shin Cheng}, {and}
  \bibinfo{person}{Ming Liu}.} \bibinfo{year}{2020}\natexlab{b}.
\newblock \showarticletitle{An Advanced LiDAR Point Cloud Sequence Coding
  Scheme for Autonomous Driving}. In \bibinfo{booktitle}{\emph{{MM} '20: The
  28th {ACM} International Conference on Multimedia}}.
  \bibinfo{publisher}{{ACM}}, \bibinfo{pages}{2793--2801}.
\newblock


\bibitem[\protect\citeauthoryear{Tian, Shen, Chen, and He}{Tian
  et~al\mbox{.}}{2019}]%
        {FCOS}
\bibfield{author}{\bibinfo{person}{Zhi Tian}, \bibinfo{person}{Chunhua Shen},
  \bibinfo{person}{Hao Chen}, {and} \bibinfo{person}{Tong He}.}
  \bibinfo{year}{2019}\natexlab{}.
\newblock \showarticletitle{{FCOS}: Fully Convolutional One-Stage Object
  Detection}. In \bibinfo{booktitle}{\emph{2019 IEEE/CVF International
  Conference on Computer Vision (ICCV)}}. \bibinfo{pages}{9626--9635}.
\newblock


\bibitem[\protect\citeauthoryear{Vora, Lang, Helou, and Beijbom}{Vora
  et~al\mbox{.}}{2020}]%
        {PointPainting}
\bibfield{author}{\bibinfo{person}{Sourabh Vora}, \bibinfo{person}{Alex~H.
  Lang}, \bibinfo{person}{Bassam Helou}, {and} \bibinfo{person}{Oscar
  Beijbom}.} \bibinfo{year}{2020}\natexlab{}.
\newblock \showarticletitle{PointPainting: Sequential Fusion for {3D} Object
  Detection}. In \bibinfo{booktitle}{\emph{2020 {IEEE/CVF} Conference on
  Computer Vision and Pattern Recognition (CVPR)}}.
  \bibinfo{pages}{4603--4611}.
\newblock


\bibitem[\protect\citeauthoryear{Wang, Fathi, Kundu, Ross, Pantofaru,
  Funkhouser, and Solomon}{Wang et~al\mbox{.}}{2020}]%
        {pillar-od}
\bibfield{author}{\bibinfo{person}{Yue Wang}, \bibinfo{person}{Alireza Fathi},
  \bibinfo{person}{Abhijit Kundu}, \bibinfo{person}{David~A. Ross},
  \bibinfo{person}{Caroline Pantofaru}, \bibinfo{person}{Thomas~A. Funkhouser},
  {and} \bibinfo{person}{Justin Solomon}.} \bibinfo{year}{2020}\natexlab{}.
\newblock \showarticletitle{Pillar-Based Object Detection for Autonomous
  Driving}. In \bibinfo{booktitle}{\emph{European Conference on Computer Vision
  (ECCV)}}, Vol.~\bibinfo{volume}{12367}. \bibinfo{pages}{18--34}.
\newblock


\bibitem[\protect\citeauthoryear{Wang, Fu, Wang, Xiao, and Dai}{Wang
  et~al\mbox{.}}{2019}]%
        {SCNet}
\bibfield{author}{\bibinfo{person}{Zhiyu Wang}, \bibinfo{person}{Hao Fu},
  \bibinfo{person}{Li Wang}, \bibinfo{person}{Liang Xiao}, {and}
  \bibinfo{person}{Bin Dai}.} \bibinfo{year}{2019}\natexlab{}.
\newblock \showarticletitle{SCNet: Subdivision Coding Network for Object
  Detection Based on {3D} Point Cloud}.
\newblock \bibinfo{journal}{\emph{{IEEE} Access}}  \bibinfo{volume}{7}
  (\bibinfo{year}{2019}), \bibinfo{pages}{120449--120462}.
\newblock


\bibitem[\protect\citeauthoryear{Wen, Han, Youk, and Liu}{Wen
  et~al\mbox{.}}{2020}]%
        {seg_pc_acmmm20}
\bibfield{author}{\bibinfo{person}{Xin Wen}, \bibinfo{person}{Zhizhong Han},
  \bibinfo{person}{Geunhyuk Youk}, {and} \bibinfo{person}{Yu{-}Shen Liu}.}
  \bibinfo{year}{2020}\natexlab{}.
\newblock \showarticletitle{{CF-SIS:} Semantic-Instance Segmentation of 3D
  Point Clouds by Context Fusion with Self-Attention}. In
  \bibinfo{booktitle}{\emph{{MM} '20: The 28th {ACM} International Conference
  on Multimedia}}. \bibinfo{publisher}{{ACM}}, \bibinfo{pages}{1661--1669}.
\newblock


\bibitem[\protect\citeauthoryear{Wu, Jiao, Yang, Zha, and Chen}{Wu
  et~al\mbox{.}}{2019}]%
        {seg_pc_acmmm19}
\bibfield{author}{\bibinfo{person}{Jian Wu}, \bibinfo{person}{Jianbo Jiao},
  \bibinfo{person}{Qingxiong Yang}, \bibinfo{person}{Zheng{-}Jun Zha}, {and}
  \bibinfo{person}{Xuejin Chen}.} \bibinfo{year}{2019}\natexlab{}.
\newblock \showarticletitle{Ground-Aware Point Cloud Semantic Segmentation for
  Autonomous Driving}. In \bibinfo{booktitle}{\emph{{MM} '19: The 27th {ACM}
  International Conference on Multimedia}}. \bibinfo{publisher}{{ACM}},
  \bibinfo{pages}{971--979}.
\newblock


\bibitem[\protect\citeauthoryear{Wu, Sahoo, and Hoi}{Wu et~al\mbox{.}}{2020b}]%
        {meta_rcnn_acmmm20}
\bibfield{author}{\bibinfo{person}{Xiongwei Wu}, \bibinfo{person}{Doyen Sahoo},
  {and} \bibinfo{person}{Steven C.~H. Hoi}.} \bibinfo{year}{2020}\natexlab{b}.
\newblock \showarticletitle{Meta-RCNN: Meta Learning for Few-Shot Object
  Detection}. In \bibinfo{booktitle}{\emph{{MM} '20: The 28th {ACM}
  International Conference on Multimedia}}. \bibinfo{publisher}{{ACM}},
  \bibinfo{pages}{1679--1687}.
\newblock


\bibitem[\protect\citeauthoryear{Wu, Pan, Chen, Long, Zhang, and Yu}{Wu
  et~al\mbox{.}}{2020a}]%
        {GNN_survey}
\bibfield{author}{\bibinfo{person}{Zonghan Wu}, \bibinfo{person}{Shirui Pan},
  \bibinfo{person}{Fengwen Chen}, \bibinfo{person}{Guodong Long},
  \bibinfo{person}{Chengqi Zhang}, {and} \bibinfo{person}{Philip~S. Yu}.}
  \bibinfo{year}{2020}\natexlab{a}.
\newblock \showarticletitle{A Comprehensive Survey on Graph Neural Networks}.
\newblock \bibinfo{journal}{\emph{IEEE Transactions on Neural Networks and
  Learning Systems}} (\bibinfo{year}{2020}), \bibinfo{pages}{1--21}.
\newblock


\bibitem[\protect\citeauthoryear{{Xie}, {Xiang}, {Yu}, {Xu}, {Yang}, {Cai}, and
  {He}}{{Xie} et~al\mbox{.}}{2020}]%
        {PIRCNN}
\bibfield{author}{\bibinfo{person}{L. {Xie}}, \bibinfo{person}{C. {Xiang}},
  \bibinfo{person}{Z. {Yu}}, \bibinfo{person}{G. {Xu}}, \bibinfo{person}{Z.
  {Yang}}, \bibinfo{person}{D. {Cai}}, {and} \bibinfo{person}{X. {He}}.}
  \bibinfo{year}{2020}\natexlab{}.
\newblock \showarticletitle{{PI-RCNN:} An Efficient Multi-Sensor {3D} Object
  Detector with Point-Based Attentive Cont-Conv Fusion Module}. In
  \bibinfo{booktitle}{\emph{2020 {AAAI} Conference on Artificial Intelligence
  (AAAI)}}. \bibinfo{pages}{12460--12467}.
\newblock


\bibitem[\protect\citeauthoryear{Yan, Mao, and Li}{Yan et~al\mbox{.}}{2018}]%
        {Second}
\bibfield{author}{\bibinfo{person}{Yan Yan}, \bibinfo{person}{Yuxing Mao},
  {and} \bibinfo{person}{Bo Li}.} \bibinfo{year}{2018}\natexlab{}.
\newblock \showarticletitle{{SECOND:} Sparsely Embedded Convolutional
  Detection}.
\newblock \bibinfo{journal}{\emph{Sensors}} \bibinfo{volume}{18},
  \bibinfo{number}{10} (\bibinfo{year}{2018}), \bibinfo{pages}{3337--3354}.
\newblock


\bibitem[\protect\citeauthoryear{Yang, Liu, Hu, Wang, and Lin}{Yang
  et~al\mbox{.}}{2019a}]%
        {Reppoints}
\bibfield{author}{\bibinfo{person}{Ze Yang}, \bibinfo{person}{Shaohui Liu},
  \bibinfo{person}{Han Hu}, \bibinfo{person}{Liwei Wang}, {and}
  \bibinfo{person}{Stephen Lin}.} \bibinfo{year}{2019}\natexlab{a}.
\newblock \showarticletitle{{RepPoints}: Point Set Representation for Object
  Detection}. In \bibinfo{booktitle}{\emph{2019 IEEE/CVF International
  Conference on Computer Vision (ICCV)}}. \bibinfo{pages}{9656--9665}.
\newblock


\bibitem[\protect\citeauthoryear{Yang, Sun, Liu, and Jia}{Yang
  et~al\mbox{.}}{2020}]%
        {3DSSD}
\bibfield{author}{\bibinfo{person}{Zetong Yang}, \bibinfo{person}{Yanan Sun},
  \bibinfo{person}{Shu Liu}, {and} \bibinfo{person}{Jiaya Jia}.}
  \bibinfo{year}{2020}\natexlab{}.
\newblock \showarticletitle{{3DSSD}: Point-Based {3D} Single Stage Object
  Detector}. In \bibinfo{booktitle}{\emph{2020 IEEE/CVF Conference on Computer
  Vision and Pattern Recognition (CVPR)}}. \bibinfo{pages}{11037--11045}.
\newblock


\bibitem[\protect\citeauthoryear{Yang, Sun, Liu, Shen, and Jia}{Yang
  et~al\mbox{.}}{2019b}]%
        {STD}
\bibfield{author}{\bibinfo{person}{Zetong Yang}, \bibinfo{person}{Yanan Sun},
  \bibinfo{person}{Shu Liu}, \bibinfo{person}{Xiaoyong Shen}, {and}
  \bibinfo{person}{Jiaya Jia}.} \bibinfo{year}{2019}\natexlab{b}.
\newblock \showarticletitle{{STD}: Sparse-to-Dense {3D} Object Detector for
  Point Cloud}. In \bibinfo{booktitle}{\emph{2019 IEEE/CVF International
  Conference on Computer Vision (ICCV)}}. \bibinfo{pages}{1951--1960}.
\newblock
\showISSN{1550-5499}


\bibitem[\protect\citeauthoryear{Ye, Xu, and Cao}{Ye et~al\mbox{.}}{2020}]%
        {HVNet}
\bibfield{author}{\bibinfo{person}{Maosheng Ye}, \bibinfo{person}{Shuangjie
  Xu}, {and} \bibinfo{person}{Tongyi Cao}.} \bibinfo{year}{2020}\natexlab{}.
\newblock \showarticletitle{{HVNet}: Hybrid Voxel Network for LiDAR Based {3D}
  Object Detection}. In \bibinfo{booktitle}{\emph{2020 IEEE/CVF Conference on
  Computer Vision and Pattern Recognition (CVPR)}}.
  \bibinfo{pages}{1628--1637}.
\newblock


\bibitem[\protect\citeauthoryear{Zhang, Chi, Yao, Lei, and Li}{Zhang
  et~al\mbox{.}}{2020}]%
        {ATSS}
\bibfield{author}{\bibinfo{person}{Shifeng Zhang}, \bibinfo{person}{Cheng Chi},
  \bibinfo{person}{Yongqiang Yao}, \bibinfo{person}{Zhen Lei}, {and}
  \bibinfo{person}{Stan~Z. Li}.} \bibinfo{year}{2020}\natexlab{}.
\newblock \showarticletitle{Bridging the Gap Between Anchor-Based and
  Anchor-Free Detection via Adaptive Training Sample Selection}. In
  \bibinfo{booktitle}{\emph{2020 IEEE/CVF Conference on Computer Vision and
  Pattern Recognition (CVPR)}}. \bibinfo{pages}{9756--9765}.
\newblock


\bibitem[\protect\citeauthoryear{Zhou, Wang, and Kr{\"{a}}henb{\"{u}}hl}{Zhou
  et~al\mbox{.}}{2019b}]%
        {Centernet}
\bibfield{author}{\bibinfo{person}{Xingyi Zhou}, \bibinfo{person}{Dequan Wang},
  {and} \bibinfo{person}{Philipp Kr{\"{a}}henb{\"{u}}hl}.}
  \bibinfo{year}{2019}\natexlab{b}.
\newblock \showarticletitle{Objects as Points}.
\newblock \bibinfo{journal}{\emph{CoRR}}  \bibinfo{volume}{abs/1904.07850}
  (\bibinfo{year}{2019}).
\newblock
\showeprint[arxiv]{1904.07850}


\bibitem[\protect\citeauthoryear{Zhou, Zhuo, and Kr{\"{a}}henb{\"{u}}hl}{Zhou
  et~al\mbox{.}}{2019c}]%
        {ExtremeNet}
\bibfield{author}{\bibinfo{person}{Xingyi Zhou}, \bibinfo{person}{Jiacheng
  Zhuo}, {and} \bibinfo{person}{Philipp Kr{\"{a}}henb{\"{u}}hl}.}
  \bibinfo{year}{2019}\natexlab{c}.
\newblock \showarticletitle{Bottom-Up Object Detection by Grouping Extreme and
  Center Points}. In \bibinfo{booktitle}{\emph{2019 IEEE/CVF Conference on
  Computer Vision and Pattern Recognition (CVPR)}}. \bibinfo{pages}{850--859}.
\newblock


\bibitem[\protect\citeauthoryear{Zhou, Sun, Zhang, Anguelov, Gao, Ouyang, Guo,
  Ngiam, and Vasudevan}{Zhou et~al\mbox{.}}{2019a}]%
        {MVF}
\bibfield{author}{\bibinfo{person}{Yin Zhou}, \bibinfo{person}{Pei Sun},
  \bibinfo{person}{Yu Zhang}, \bibinfo{person}{Dragomir Anguelov},
  \bibinfo{person}{Jiyang Gao}, \bibinfo{person}{Tom Ouyang},
  \bibinfo{person}{James Guo}, \bibinfo{person}{Jiquan Ngiam}, {and}
  \bibinfo{person}{Vijay Vasudevan}.} \bibinfo{year}{2019}\natexlab{a}.
\newblock \showarticletitle{End-to-End Multi-View Fusion for 3D Object
  Detection in LiDAR Point Clouds}. In \bibinfo{booktitle}{\emph{2019 Annual
  Conference on Robot Learning (CoRL)}}, Vol.~\bibinfo{volume}{100}.
  \bibinfo{publisher}{{PMLR}}, \bibinfo{pages}{923--932}.
\newblock


\bibitem[\protect\citeauthoryear{Zhou and Tuzel}{Zhou and Tuzel}{2018}]%
        {VoxelNet}
\bibfield{author}{\bibinfo{person}{Yin Zhou} {and} \bibinfo{person}{Oncel
  Tuzel}.} \bibinfo{year}{2018}\natexlab{}.
\newblock \showarticletitle{{VoxelNet}: End-to-End Learning for Point Cloud
  Based {3D} Object Detection}. In \bibinfo{booktitle}{\emph{2018 IEEE/CVF
  Conference on Computer Vision and Pattern Recognition (CVPR)}}.
  \bibinfo{pages}{4490--4499}.
\newblock
\showISSN{1063-6919}


\bibitem[\protect\citeauthoryear{Zhu, Hu, Lin, and Dai}{Zhu
  et~al\mbox{.}}{2019}]%
        {MDCN}
\bibfield{author}{\bibinfo{person}{Xizhou Zhu}, \bibinfo{person}{Han Hu},
  \bibinfo{person}{Stephen Lin}, {and} \bibinfo{person}{Jifeng Dai}.}
  \bibinfo{year}{2019}\natexlab{}.
\newblock \showarticletitle{{Deformable ConvNets V2}: More Deformable, Better
  Results}. In \bibinfo{booktitle}{\emph{2019 IEEE/CVF Conference on Computer
  Vision and Pattern Recognition (CVPR)}}. \bibinfo{pages}{9300--9308}.
\newblock


\end{thebibliography}



\end{document}